\documentclass{article}
\usepackage{neurips_2023_preprint}

\usepackage[utf8]{inputenc} 
\usepackage[T1]{fontenc}    
\usepackage[hidelinks]{hyperref}       
\usepackage{url}            
\usepackage{booktabs}       
\usepackage{amsfonts}       
\usepackage{nicefrac}       
\usepackage{microtype}      
\usepackage{xcolor}         

\usepackage{thm-restate}
\usepackage{amsmath}
\usepackage{amssymb}
\usepackage{dsfont}
\usepackage{graphicx}
\usepackage[numbers]{natbib}
\usepackage{cleveref}
\usepackage{xspace}
\usepackage{enumitem}
\usepackage{amsmath,centernot}

\newtheorem{theorem}{Theorem}
\newtheorem{corollary}{Corollary}
\newtheorem{lemma}{Lemma}
\newtheorem{proposition}{Proposition}
\newtheorem{definition}{Definition}
\newtheorem{remark}{Remark}
\newtheorem{example}{Example}
\newtheorem{assumption}{Assumption}

\crefname{equation}{Eq.}{Eqs.}
\Crefname{equation}{Equation}{Equations}

\crefname{proposition}{Prop.}{Props.}
\Crefname{proposition}{Proposition}{Propositions}

\crefname{theorem}{Th.}{Ths.}
\Crefname{theorem}{Theorem}{Theorems}

\crefname{lemma}{Lem.}{Lems.}
\Crefname{lemma}{Lemma}{Lemmas}

\crefname{definition}{Def.}{Defs.}
\Crefname{definition}{Definition}{Definitions}

\crefname{remark}{Rem.}{Rems.}
\Crefname{remark}{Remark}{Remarks}

\crefname{corollary}{Cor.}{Cors.}
\Crefname{corollary}{Corollary}{Corollaries}

\crefname{assumption}{Asm.}{Asms.}
\Crefname{assumption}{Assumption}{Assumptions}

\crefname{section}{Sec.}{Secs.}
\Crefname{section}{Section}{Sections}

\crefname{appendix}{App.}{Apps.}
\Crefname{appendix}{Appendix}{Appendices}

\crefname{algorithm}{Alg.}{Algs.}
\Crefname{algorithm}{Algorithm}{Algorithms}

\newenvironment{proof}{\par\noindent{\bf Proof\ }}

\title{From Mutual Information to Expected Dynamics:\\ New Generalization Bounds for Heavy-Tailed SGD}

\author{Benjamin Dupuis$^1$ \and Paul Viallard$^1$}

\date{
  $^1$ Inria, CNRS, Ecole Normale Supérieure\\
  PSL Research University, Paris, France\\
  \texttt{\{benjamin.dupuis,paul.viallard\}@inria.fr}
} 

\newcommand{\betatail}{\beta_{\text{\normalfont{tail}}}}
\newcommand{\resp}{{\it resp.}\xspace}
\newcommand{\eg}{{\it e.g.}\xspace}
\newcommand{\aka}{{\it a.k.a.}\xspace}
\newcommand{\wrt}{{\it w.r.t.}\xspace}
\newcommand{\iid}{{\it i.i.d.}\xspace}
\newcommand{\Zcal}{\mathcal{Z}}
\newcommand{\ie}{{\it i.e.}\xspace}

\newcommand{\defeq}{:=}
\newcommand{\R}{\mathds{R}}
\newcommand{\risk}{\mathcal{R}}
\newcommand{\er}{\widehat{\mathcal{R}}_S}
\newcommand{\totalmutual}{\text{I}_\infty}

\newcommand{\zcal}{\mathcal{Z}}
\newcommand{\wcal}{\mathcal{W}}
\newcommand{\wcalsu}{\mathcal{W}_{S,U}}
\newcommand{\ycal}{\mathcal{Y}}
\newcommand{\ycalu}{\mathcal{Y}_{U}}
\newcommand{\wgen}{\sup_{w\in\wcalsu} \left( \risk(w) - \er(w) \right)}

\newcommand{\Rd}{\R^d}
\newcommand{\levy}{L_t^\alpha}
\newcommand{\upperbox}{\overline{\dim}_{\text{\normalfont{B}}}}
\newcommand{\gnabla}{ \sup_{w \in \ycalu} \Vert \nabla\er(w) - \nabla \risk(w) \Vert}
\newcommand{\E}{\mathds{E}}
\newcommand{\Eof}[2][]{\mathds{E}_{#1} \left[ #2 \right]}
\newcommand{\Pof}[2][]{\mathds{P}_{#1} \left( #2 \right)}

\newcommand{\symb}{\boldsymbol{\Psi}}
\newcommand{\prob}{\mathds{P}}
\newcommand{\normof}[1]{\left\Vert #1 \right\Vert}
\newcommand{\klb}[2]{\text{\normalfont{{KL}}}\left(#1 \right|\left| #2 \right)}
\newcommand{\renyi}[3][\alpha]{\text{\normalfont{{D}}}_{#1}\left(#2 \right|\left| #3 \right)}

\newcommand{\hausdim}{\dim_{\text{\normalfont{H}}}}
\newcommand{\sgn}{\text{\normalfont{sgn}}}
\newcommand{\support}{\text{\normalfont{supp}}}
\newcommand{\nll}{\centernot{\ll}}
\newcommand{\multiindex}{{\boldsymbol{i}}}
\newcommand{\partialmean}[1][j]{\overline{\partial_{#1}}}
\newcommand{\intr}{\int_{-\infty}^\infty}
\newcommand{\phieps}{\varphi_\epsilon}
\DeclareMathOperator*{\esssup}{ess\,sup}
\newcommand{\dimvalue}{\gamma}

\begin{document}

\maketitle

\begin{abstract}
Understanding the generalization abilities of modern machine learning algorithms has been a major research topic over the past decades. In recent years, the learning dynamics of Stochastic Gradient Descent (SGD) have been related to heavy-tailed dynamics. This has been successfully applied to generalization theory by exploiting the fractal properties of those dynamics. However, the derived bounds depend on mutual information (decoupling) terms that are beyond the reach of computability. In this work, we prove generalization bounds over the trajectory of a class of heavy-tailed dynamics, without those mutual information terms. Instead, we introduce a geometric decoupling term by comparing the learning dynamics (depending on the empirical risk) with an expected one (depending on the population risk). We further upper-bound this geometric term, by using techniques from the heavy-tailed and the fractal literature, making it fully computable. Moreover, as an attempt to tighten the bounds, we propose a PAC-Bayesian setting based on perturbed dynamics, in which the same geometric term plays a crucial role and can still be bounded using the techniques described above.
\end{abstract}

\section{Introduction}

In machine learning, we aim to solve a learning task represented by an unknown probability distribution $\mu_z$ over a set of data probability space $(\Zcal, \mathcal{F})$, where $\mathcal{F}$ is the associated $\sigma$-algebra.
In order to solve it, we aim to learn a machine learning model parameterized by some weights $w\in\R^d$.
For instance, the model can be obtained by solving the \emph{Population Risk Minimization} problem defined by
\begin{align}
\min_{w\in \Rd} \Big\{ \risk(w) \defeq \mathds{E}_{z\sim\mu}\ell(w, z) \Big\}, \tag{PRM}
\end{align}
where $\ell: \Rd \times \mathcal{Z} \to \R$ is the loss function given the weights,  $z\in\zcal$ an example and $\risk(w)$ the population risk.
However, since the probability distribution $\mu_z$ is unknown, we cannot directly solve PRM to obtain a model.
Instead, we have at our disposal $n$ data points $S=\{z_i\}_{i=1}^{n}$, sampled \iid from $\mu_z$ that serve to obtain a model.
Thanks to the data, a model can be obtained by (approximately) solving the \emph{Empirical Risk Minimization} problem defined as
\begin{align*}
\min_{w\in \Rd} \left\{ \er(w) \defeq {\textstyle\frac{1}{n}\sum_{i=1}^n\ell(w, z_i)}\right\}, \tag{ERM}
\end{align*}
where $\er(\cdot)$ is the \emph{empirical risk}.
Nowadays, one workhorse algorithm used to solve the ERM problem is the Stochastic Gradient Descent (SGD) algorithm.
Starting from weights $W_0\in\Rd$, the algorithm updates the weights iteratively by applying the following update at each time $t$
\begin{align*}
W_{t+1} = W_t - \eta\nabla \widehat{\mathcal{R}_B}(W_t),
\end{align*}
where $\eta>0$ is the learning rate and $B\subseteq S$ is a random subset of indices.
Given $B$ and $\eta$, there is theoretical and empirical evidence that the gradient noise $\nabla\widehat{\mathcal{R}_B}(W_t){-}\nabla\widehat{\mathcal{R}_S}(W_t)$ can be \emph{heavy-tailed} \citep{simsekli_tail-index_2019,gurbuzbalaban_heavy-tail_2021,hodgkinson_multiplicative_2021,wang_convergence_2021}, \ie, there is a high probability that the gradient noise is large.
One way of modeling SGD with such a gradient noise can be done thanks to the following stochastic differential equation (SDE) \citep{nguyen_first_2019,raj_algorithmic_2023,simsekli_hausdorff_2021,simsekli_tail-index_2019}
\begin{align}
    \label{eq:empirical_dynamics_intro}
    dW_t = -\nabla\er(W_t)dt + dL_t^\alpha,
\end{align}
where $L_t^\alpha$ is a $d$-dimensional $\alpha$-stable Lévy process (defined more formally further), where $\alpha$ is the ``tail-index''.
Under this modeling assumption, \emph{generalization bounds} (see \citet{mohri_foundations_2012}), \ie, upper bounds on the generalization gap $\risk(w)-\er(w)$ have been derived.
More precisely, most works focus on the \emph{worst-case generalization error} over the trajectory.
\citet[Thms. 1 and 2]{simsekli_hausdorff_2021}, \citet[Thm 1. with Cor. 1]{hodgkinson_generalization_2021}, and \citet[Prop. 1]{birdal_intrinsic_2021} propose generalization bounds that have informally the following form, with probability at least $1{-}\zeta$
\begin{align}
    \label{eq:fractal_bounds_informal}
    \sup_{w \in \mathcal{W}_S} \left( \risk(w) - \er(w) \right) \lesssim \sqrt{\frac{\upperbox + \log(1/\zeta) + \totalmutual(\mathcal{W_S}, S)}{n}},
\end{align}
where, $\wcal_S := \{ W_t, ~0\leq t \leq T \}$ is the trajectory of the dynamics up to a certain time horizon, $\totalmutual(\wcal_S, S)$ is a total mutual information term, quantifying the statistical dependence between $S$ and $\wcal_S$, see \citep[Section $2.2$]{hodgkinson_generalization_2022}, and $\upperbox$ is a notion of fractal dimension\footnote{We refer the reader to \citep{falconer_fractal_2014} for an introduction to fractal geometry.}, which can be seen as a measure of the complexity of $\wcal_S$ (defined formally in \cref{sec:fractal_geometry}). 
The fractal geometry of those stochastic processes, like \cref{eq:fractal_bounds_informal}, has been extensively studied \citep{xiao_random_2004} and provided powerful tools to study the generalization properties of such heavy-tailed dynamics.
\citet{simsekli_hausdorff_2021} noted that, under reasonable assumptions, the dimension $\upperbox$, appearing in \cref{eq:fractal_bounds_informal}, is equal to the corresponding \emph{Hausdorff dimension} and may be related to the tail-index $\alpha$.

Even if the fractal dimension could be approximated by estimating $\alpha$, or by using the method developed by \citet{birdal_intrinsic_2021}, the total mutual information term appearing in the bound is not computable, making the bound beyond the reach of any evaluation.
Moreover, the total mutual information can be large or even infinite, making the bound potentially vacuous.
To avoid this inconvenience, \citet{raj_algorithmic_2023} proposed bounds for \cref{eq:empirical_dynamics_intro} without mutual information terms based on a stability approach.
Their method is based on comparing two dynamics, using a surrogate loss, associated with two datasets differing by one point.
However, their bounds are bounds in expectation over the last point of the dynamics and have an important dependence on the number of parameters $d$, while we bound the worst-case generalization gap, over the trajectory, with high probability. 

In this work, we aim to close this gap by proposing in \cref{sec:main-result-sup} a new upper bound on $\sup_{w \in \mathcal{W}_S} \risk(w) - \er(w)$ using purely geometric and computable quantity, instead of the total mutual information. 
This is done by the comparison of the SDE in \cref{eq:empirical_dynamics_intro} with the corresponding \emph{expected dynamics}, where we consider the population risk in the dynamic instead of the empirical one. 
Most importantly, we further bound the introduced geometric term using properties of the heavy-tailed dynamics.
As an attempt to obtain a tighter bound, in \cref{sec:main-result-pac-bayes}, we present a PAC-Bayesian setting in which the aforementioned geometric terms naturally appear, and we prove new bounds on the generalization gap $\risk(w){-}\er(w)$, based on this PAC-Bayesian framework. 
In this case, the big constants appearing in the bounds of \cref{sec:main-result-sup}, which in particular are exponential in time, are compensated by a higher rate in $n$ (\eg, linear), therefore substantially improving the bounds.
We also highlight in \cref{sec:additional_results} that, under additional assumptions, time independence may be obtained in this PAC-Bayesian setting.
Note that all proofs are deferred in the appendices.

\section{Setting and Notations}

\subsection{Learning Dynamics}

Let $(\Omega, \mathcal{T}, \mu_u)$ be a complete probability space, representing the external randomness added by the algorithms. 
To match with notations in \citep{simsekli_hausdorff_2021,dupuis_generalization_2023}, the randomness will generally be denoted $U$.
We consider, on $\Omega$, the two dynamics, in $\Rd$, that we call respectively \emph{empirical} and \emph{expected} dynamics:
\begin{align}
dW^S_t = -\nabla \er(W^S_t)dt + dL_t^\alpha,& \quad\text{with}\quad W_0 = Z_0,\label{eq:empirical_dynamics}\\
\text{and}\quad dY_t = -\nabla \risk(Y_t)dt + dL_t^\alpha,& \quad\text{with}\quad Y_0 = Z_0,\label{eq:expected_dynamics}
\end{align}

where $Z_0$ is a random weight independent of all other random variables. 
Recall that $\levy$ is a strictly $\alpha$-stable Lévy process; see \cref{sec:levy_feller}, or the tutorial \citep{schilling_introduction_2016}, for formal definitions.
Note that \cref{eq:empirical_dynamics} is similar to models considered in other works \citep{simsekli_tail-index_2019,nguyen_first_2019,raj_algorithmic_2023} and is a particular case of \citep[Eq. $4$]{simsekli_hausdorff_2021}. 
To ensure that the SDEs in \cref{eq:empirical_dynamics,eq:expected_dynamics} have strong solutions and are Feller processes (see \cref{thm:sde_existence_solution}), the following assumption imposes that the terms $-\nabla \risk(\cdot)$, $-\nabla \er(\cdot)$ are smooth enough.
\begin{assumption}
    \label{ass:smooth_lipschitz}
    The loss is differentiable, $L$-Lipschitz continuous and $M$-smooth, \ie, for all $w, w'\in \Rd$ and $z \in \zcal$, we have:
    \begin{align*}
    |\ell(w,z) - \ell(w',z) | \leq L \Vert w - w' \Vert,\quad \text{and} \quad
\Vert \nabla \ell(w,z) - \nabla\ell(w',z) \Vert \leq M \Vert w - w' \Vert.
    \end{align*}
\end{assumption}

This assumption plays a great role in the generalization bounds presented in \cref{sec:main-result-sup,sec:main-result-pac-bayes}.
As it is common to derive generalization bounds (see \eg, \citep{xu_information-theoretic_2017,simsekli_hausdorff_2021}), we make a sub-Gaussian assumption on the distribution of $\ell(w,z)$, to control its concentration around its expectation.
\begin{assumption}
    \label{ass:sub_Gaussian}
    The function $\ell: \Rd \times \zcal \longrightarrow \mathds{R}$ is assumed to be $\sigma$-sub-Gaussian, \ie, for all $\lambda > 0$, we have $\mathds{E} \left[ e^{\lambda (\ell(w, z) - \mathds{E}[\ell(w, z)])}  \right] \leq e^{\sigma^2 \lambda^2/2}$.
\end{assumption}

In this paper, we also fix the time horizon $T>0$ of the processes that we denote by $Y_t(U)$ and $W^S_t(U)$ by omitting the variable $U$ when it is not ambiguous.
Let us also introduce the trajectories $\wcalsu$ and $\ycal_{U}$ associated with the processes $Y_t(U)$ and $W^S_t(U)$ and defined by
\begin{align*}
\wcalsu = \{ W^S_t(U),~0\leq t \leq T\}   ,\quad \ycal_{U} =\{ Y_t(U),~0\leq t \leq T\},
\end{align*}
which is the sets of weights $W^S_t(U)$ and $Y_t(U)$ encountered by the processes.

\begin{remark}
    \label{rq:well-posedness}
    \cref{eq:empirical_dynamics} is here understood as holding at all times, almost surely with respect to $\mu_z^{\otimes n} \otimes \mu_u$.
    Moreover, we assume that the map $(S,U,t)\longmapsto W^S_t(U)$ is (jointly) measurable. In particular, this assumption can be true if $\zcal$ is countable.
    We also implicitely use a càdlàg version of $\levy$, without loss of generality \citep[Section $2$]{schilling_introduction_2016}, \citep[Theorem $2.9$]{revuz_continuous_1999}.
\end{remark}

In addition to \cref{ass:smooth_lipschitz}, we make the following topological assumption on the trajectory of $\ycal$, which will help us control its fractal properties and relate them to the tail-index $\alpha$, it is similar to assumptions already made in the literature, see \citep[Assumption H$4$]{simsekli_hausdorff_2021}. 

\begin{assumption}[Fractal characterization of the dynamics]
    \label{ass:ahlfors_y}
    We assume that, almost surely, the Hausdorff and upper box-counting dimension of $\ycalu$ are the same, \ie, $\upperbox(\ycalu) = \hausdim(\ycalu)$. In all the following, this common dimension value will be denoted $\gamma$. Note that we provide technical background, regarding fractal geometry, in \cref{sec:fractal_geometry}.
\end{assumption}

Roughly speaking, we implicitly assume that the process $(Y_t)_{t \geq 0}$ is ``regular'' enough (which happens for smooth dynamics).
More formally, this is particularly true if $\alpha=2$ or if we assume the so-called Ahlfors regularity of $\ycalu$ (\citep[Assumption H$4$]{simsekli_hausdorff_2021} and \citep[Corollary $1$]{hodgkinson_generalization_2022}), see \cref{sec:fractal_geometry}.

\subsection{Heavy-tailed Properties}

In our particular setting, we can conduct a precise analysis of the relation between the Hausdorff dimension of $\ycal$ and $\wcalsu$ and the tail-index $\alpha$. 
The following theorem describes some relationships between this dimension and the tail-index $\alpha$, for the dynamics considered in this paper. 
It is a direct consequence of existing works on the fractal dimensions of Feller processes \citep{schilling_feller_1998,bottcher_levy_2013,knopova_lower_2015}.

\begin{restatable}[Fractal properties of heavy-tailed dynamics (informal)]{theorem}{thmFractalProperties}
    \label{thm:heavy_tiled_sdes}
    Under \cref{ass:sub_Gaussian,ass:smooth_lipschitz}, if one of the following condition is true: {\it (i)} $\alpha \in [1,2]$, or {\it (ii)} $L_t^\alpha$ is strictly stable and $\nabla \ell$ is smooth enough in $w$ (compact support is enough for instance) so that $Y$ and $W^S$ satisfy the assumptions of \citep[Theorem $4$]{schilling_feller_1998}, then $\hausdim(\ycalu), \hausdim(\wcalsu) \leq  \alpha$.
    Hence, under \cref{ass:ahlfors_y}, we have $\gamma \leq \alpha$.
\end{restatable}

In the rest of the paper, to avoid overloading the statements with technical assumptions, we will keep the notations $\gamma$ to refer to the fractal dimension of the trajectories.
However, $\gamma$ may be replaced by $\alpha$, in our upper bounds, whenever \cref{ass:ahlfors_y} and \cref{thm:heavy_tiled_sdes} hold.

\section{From Statistical to Geometric Information}
\label{sec:main-result-sup}

In this section, we present a generalization bound, in \cref{thm:lipschitz_master_theorem}, on the worst-case generalization gap in terms of the fractal dimension $\gamma$ of the trajectory, but without mutual information term, compared to \citep{simsekli_hausdorff_2021,birdal_intrinsic_2021,hodgkinson_generalization_2022,dupuis_generalization_2023}. 
The following theorem introduces a Euclidean distance between the empirical and expected trajectories and uses \citep[Lemma $1$]{simsekli_hausdorff_2021} to bound the generalization error over the data-dependent trajectory $\wcalsu$.

\begin{restatable}{theorem}{theoremhausdorff}
\label{thm:lipschitz_master_theorem}
    Fix some $\zeta \in (0,1)$. Under \cref{ass:sub_Gaussian,ass:smooth_lipschitz,ass:ahlfors_y}, there exists $N \in \mathds{N}$, depending only on $\zeta$ such that, with probability at least $1 - 2\zeta$ over $\mu_u \otimes \mu^{\otimes n}$, for all $n\geq N$, we have:
    \begin{align*}
        \sup_{w \in \mathcal{W}_{S,U}} \left( \risk(w) - \er(w) \right) \leq 2L \sup_{0\leq t \leq T} \Vert W^S_t - Y_t \Vert   + \frac{2}{\sqrt{n}}  + 2\sigma \sqrt{\frac{2\log(L\sqrt{n}) \dimvalue + \log(1/\zeta)}{2n}}.
    \end{align*}
\end{restatable}

\begin{remark}
    \cref{thm:lipschitz_master_theorem} would still be true if we replace the term $\sup_{0\leq t \leq T} \Vert W^S_t - Y_t \Vert$ by the Hausdorff distance $d_H(\wcalsu, \ycalu)$ between $\wcalsu$ and $\ycal$, see the proof for more details.
\end{remark}

Introducing the geometric term $\sup_{0\leq t \leq T} \Vert W^S_t - Y_t \Vert$ allows us to transfer the estimation of the generalization error from $\wcalsu$ to $\ycalu$ (which does not depend on the data).
Note that it is not clear, when looking at \cref{thm:lipschitz_master_theorem}, that the given bound is non-vacuous, as the distance between the trajectories is not divided by $\sqrt{n}$. 
However, it is intuitive to think that $\sup_{0\leq t \leq T} \Vert W^S_t - Y_t \Vert$ could decrease rapidly as $n \to \infty$, because of the convergence of $\er(\cdot)$ to $\risk(\cdot)$.
Therefore, in the rest of this section, we will derive bounds on this term, hence making our generalization bound more pertinent.
An important property, of our strategy to achieve this goal, is that we consider \cref{ass:smooth_lipschitz} and then leverage the fractal properties of the heavy-tailed trajectories by performing a covering argument and introducing the fractal dimension, see \cref{sec:posponed_proof} for more details. 
The aforementioned covering argument is used to bound the worst-case concentration of the gradients $\nabla \ell(w,z)$, whose norm is sub-Gaussian thanks to the Lipschitz loss assumption. 
The drawback of this approach is to yield a big dependence of the bound on the Lipschitz constant $L$.
However, in the best-case scenario where the distribution of the gradient norm is evenly shared among its components, we note that it would mean that each partial derivative $\tfrac{\partial \ell(w,z)}{\partial w_j}$ has a sub-Gaussian norm \citep[Definition $2.5.6$]{vershynin_high-dimensional_2020} of order $\mathcal{O}(1/\sqrt{d})$.
Thus, we argue that a strong concentration of those derivatives can lead to a better bound, and, \emph{when specified}, we shall make the following assumption.

\begin{assumption}[Partial derivatives concentration]
    \label{ass:partal_derivatives_concentration}
    For $1 \leq j \leq d$, the random variable $\partial_j \ell(w,z)$ is $\Sigma/\sqrt{d}$-sub-Gaussian, uniformly on $j$ and $w$, where $\Sigma > 0$ is a constant. 
\end{assumption}

The following theorem provides a bound on the geometric term, $\sup_{0\leq t \leq T} \Vert W^S_t - Y_t \Vert$, using the methods described above under different sets of assumptions.
\begin{restatable}{theorem}{theoremBoundingHausdorff}
    \label{thm:second_main_result}
    Let us fix some $\zeta \in (0,1)$. Under \cref{ass:sub_Gaussian,ass:smooth_lipschitz,ass:ahlfors_y}, there exists $N \in \mathds{N}$, depending on $\zeta$, such that, with probability at least $1 - 2\zeta$ over $\mu_u \otimes \mu^{\otimes n}$, for all $n\geq N$, we have:
    \begin{align*}
        \sup_{0\leq t \leq T} \Vert W^S_t - Y_t \Vert \leq \frac{e^{MT} - 1}{M} \left( \frac{2}{\sqrt{n}} + L \sqrt{\frac{2}{n}} + 2L \sqrt{\frac{2\dimvalue \log(M\sqrt{n}) + \log(1/\zeta)}{2n}} \right).
    \end{align*}
    If we further make \cref{ass:partal_derivatives_concentration}, we can improve the constants to:
    \begin{align*}
        \sup_{0\leq t \leq T} \Vert W^S_t - Y_t \Vert \leq \frac{e^{MT} - 1}{M} \left( \frac{2}{\sqrt{n}} +  2 \Sigma \sqrt{\frac{2\dimvalue \log(M\sqrt{n}) + \log(2d/\zeta)}{2n}} \right).
    \end{align*}    
\end{restatable}

Therefore, the tail-index $\alpha$ (upper-bounding $\gamma$) also has an impact on $\sup_{0\leq t \leq T} \Vert W^S_t - Y_t \Vert$. 
While the bounds of \cref{thm:second_main_result} may be made vacuous because of the dependence on $T$ and the Lipschitz constant $L$, we highlight that it has the advantage of being explicit, as opposed to all existing worst-case heavy-tailed generalization bounds in high probability. 
In some sense, we replaced the statistical decoupling (based on mutual information terms) with a new kind of geometric term, based on the Euclidean distance between the empirical and the expected dynamics.
Therefore, by combining \cref{thm:lipschitz_master_theorem,thm:second_main_result}, we obtain the following corollary.

\begin{restatable}{theorem}{theoremHausdorffBoundGenBound}
    \label{thm:second_main_result_corollary}
    Let us fix some $\zeta \in (0,1)$. Under \cref{ass:sub_Gaussian,ass:smooth_lipschitz,ass:ahlfors_y}, there exists $N \in \mathds{N}$, depending only on $\zeta$ such that, with probability at least $1 - 4\zeta$ over $\mu_u \otimes \mu^{\otimes n}$, for all $n\geq N$, we have:
    \begin{align*}
        \sup_{w \in \mathcal{W}_{S,U}} \left( \risk(w) - \er(w) \right) \leq \frac{2}{\sqrt{n}} + \frac{2L}{\sqrt{n}}C_1\frac{e^{MT} - 1}{M} +  C_2 \sqrt{\frac{2\log(C_0\sqrt{n}) \dimvalue + \log(C_3/\zeta)}{2n}},
    \end{align*}
    with constants 
    \begin{align*}
    C_0 \defeq \max\{M{,}L\},\ \ C_1 \defeq 2 + \sqrt{2}L, \ \ C_2 \defeq 4L^2\tfrac{1}{M}(e^{MT}{-}1) + 2\sigma, \ \ \text{and}\ \ C_3 = 1.
    \end{align*}
    If we further make \cref{ass:partal_derivatives_concentration}, we can improve the constants to (with a little loss on $C_3$) to 
    \begin{align*}
    C_1 \defeq 2, \quad C_2 \defeq 4\Sigma\tfrac{L}{M}(e^{MT}{-}1) + 2\sigma, \quad C_3 = 2d.
    \end{align*}
\end{restatable}

\cref{thm:second_main_result_corollary} corresponds to a direct way of using expected dynamics, at the cost of introducing huge constants in the derived bounds. In the next subsection, we propose a first step toward a better understanding of the geometric term $\sup_{0\leq t \leq T} \Vert W^S_t - Y_t \Vert$, based on the PAC-Bayesian theory.

\section{Tightening the Bounds Through the PAC-Bayesian Theory}
\label{sec:main-result-pac-bayes}

While the method described in the previous section allows us to prove bounds for heavy-tailed dynamics without any mutual information term, it makes the bound rely too much on the Lipschitz continuity assumption, and has an exponential dependence on the training time, which may render the bound vacuous.
In this section, we argue that the comparison between empirical and expected dynamics is able to induce better bounds. 
Inspired by PAC-Bayesian theory\footnote{See \citep{alquier_user-friendly_2021,hellstrom_generalization_2023} for an introduction to the PAC-Bayesian theory.} \citep{shawetaylor_pac_1997,mcallester_some_1998}, we want to prove bounds where the supremum over the trajectory $\sup_{w \in \mathcal{W}_{S,U}}(\risk(w){-}\er(w))$ is replaced by a uniform sampling over the trajectory. 
To do so, we define a distribution for the trajectory $\mathcal{W}_{S,U}$:
\begin{align}
\label{eq:unperturbd_priors_posteriors}
    \forall B \in \mathcal{B}(\Rd),\quad \Tilde{\rho}_{S,U}(B) := \frac{1}{T} \int_0^T \mathds{1} \left( W_t^S(U) \in B \right) dt,
\end{align}
where $\mathcal{B}(\Rd)$ is the Borel $\sigma$-algebra of $\Rd$ and $\mathds{1}(\cdot)$ is the indicator function.
This distribution can be interpreted as the distribution of the weights that are sampled uniformly over time.
It is related to the notion of occupation measure in stochastic calculus (see \citep[Section $6$]{xiao_random_2004}).

\begin{remark}
Within our fractal framework, the distribution $\Tilde{\rho}_{S,U}$ can be replaced with the exact Hausdorff measure restricted on the trajectories, assuming those measures are finite.
However, it appears in several works, like \citep[Theorem $5.1$]{xiao_random_2004}, that those measures are often, up to multiplicative constants, equivalent to the occupation measures \citep[Section $6$]{xiao_random_2004} of the processes, making the sampling in time and the sampling of the points similar.
\end{remark}

PAC-Bayesian analysis requires the comparison between the data-dependent distribution $\Tilde{\rho}_{S,U}$ with a data-independent distribution, \ie, a prior distribution. 
In order to leverage the comparison between the empirical dynamics \eqref{eq:empirical_dynamics} and the expected one \eqref{eq:expected_dynamics}, the natural choice is to define the prior (denoted $\Tilde{\pi_U}$) in the same way than $\Tilde{\rho}_{S,U}$, but with support on the expected trajectory, \ie, $\Tilde{\pi_U}(B) = (1/T)\int_0^T \mathds{1} \left( Y_t(U) \in B \right) dt$.
While this choice is appealing, it has a major drawback. 
Indeed, those distributions do not satisfy the absolute continuity condition required by the PAC-Bayesian framework (\ie, $\Tilde{\rho}_{S,U} \nll \Tilde{\pi}_{U}$). In order to overcome this issue, we chose to perturb the distribution $\Tilde{\rho}_{S,U}$ by a Gaussian one.
This leads to the following definitions of the distribution on both trajectories $\mathcal{W}_{S,U}$:
\begin{align}
    \label{eq:perturbed_priors_posteriors}
    \rho_{S,U} \defeq \mathcal{N}(0,s^2) * \Tilde{\rho}_{S,U}, \quad \pi_{U} \defeq \mathcal{N}(0,s^2) * \Tilde{\pi}_{U}
\end{align} 
where $*$ denotes the convolution.
This kind of smoothed distribution has already been considered by \citet{lugosi_generalization_2022}, but not for trajectories, like in \cref{eq:perturbed_priors_posteriors}.
We will call $\rho_{S,U}$ the posterior distributions and $\pi_U$ the prior distributions.
Informally, sampling from $\rho_{S,U}$ means sampling points that are not too far from the true trajectory $\wcalsu$, as if we were studying a learning algorithm approximating, in some sense, those continuous trajectories.
The scale parameter $s$ then controls the quality of this approximation.

We present two results (respectively in \cref{thm:pac_bayes_wassertein} and \cref{thm:disintegrated_heavy_tailed_bound}): {\it (a)} a bound in expectation over $\rho_{S,U}$, like in the ``classical'' PAC-Bayesian theory (see \eg, \citep{germain_pac-bayesian_2009}) and {\it (b)} a bound in high-probability over $\rho_{S,U}$ with the ``disintegrated'' PAC-Bayesian theory~\citep{viallard_general_2021}. 
The power of our approach is that we are able to bound the information-theoretic terms, \ie, the KL (\resp Rényi) divergence $\klb{\rho_{S,U}}{\pi_U}$ (\resp $\renyi[\beta]{\rho_{S,U}}{\pi_U}$), appearing in those bounds by the same geometric term than in \cref{sec:main-result-sup}, see \cref{sec:proof_of_pac_bayes_bound} for details.
Therefore, the analysis of \cref{sec:main-result-sup} still applies.
The next theorem presents our first bound, which is an application of \cref{thm:kl_pac_bayes__bound} in our setting.

\begin{restatable}{theorem}{theoremPACBayesWassertein}
    \label{thm:pac_bayes_wassertein}
    Under \cref{ass:sub_Gaussian,ass:smooth_lipschitz}, for any $\lambda{>}0$, with probability at least $1{-}\zeta$ over $(S{,}U)$ we have:
    \begin{align*}
      \lambda\mathds{E}_{w \sim \rho_{S,U}}[\risk(w) - \er(w)] \leq \frac{1}{Ts^2}  \int_0^T \Vert W_u^S - Y_u \Vert^2 du+ \log(1/\zeta) + \frac{C \lambda^2}{n}.
    \end{align*}
\end{restatable}
Note that we can further bound the integral term in \cref{thm:pac_bayes_wassertein} with the same term than in \cref{thm:lipschitz_master_theorem}, \ie,
\begin{align*}
\frac{1}{T} \int_0^T \Vert W_u^S - Y_u \Vert^2 du \leq   \sup_{0\leq t \leq T} \Vert W^S_t - Y_t \Vert^2.
\end{align*}
We now state the corresponding disintegrated bound, using the framework developed by \citet{viallard_general_2021}, in which we obtain the same geometric distances as in \cref{thm:lipschitz_master_theorem,thm:pac_bayes_wassertein}.

\begin{restatable}{theorem}{thmDisintegrated}
    \label{thm:disintegrated_heavy_tailed_bound}
    Under \cref{ass:sub_Gaussian,ass:smooth_lipschitz}, for any $\lambda > 0$ and $\beta > 1$, we have, with probability at least $1 - \zeta$ over the joint distribution of $(S,U)$ and $w\sim\rho_{S,U}$:
    \begin{align*}
        \frac{\lambda \beta}{\beta - 1} \left( \risk(w) - \er(w) \right) \leq \frac{2\beta - 1}{\beta - 1} \log \left( \frac{2}{\zeta} \right) + \frac{\beta}{2s^2} \sup_{0\leq t \leq T} \Vert W^S_t - Y_t \Vert^2 + \left( \frac{\beta}{\beta - 1} \right)^2 \frac{\lambda^2 \sigma^2}{2n}.
    \end{align*}
\end{restatable}

\cref{thm:pac_bayes_wassertein} provides a bound by integrating the generalization error over $\rho_{S,U}$, which can be seen as a mean error near the empirical trajectory $\wcalsu$.
\cref{thm:disintegrated_heavy_tailed_bound} provides more precise information, it gives a high probability bound by sampling weights near the empirical trajectory, this comes at the cost of replacing the integral distance, \ie, $\frac{1}{T} \int_0^T \Vert W_u^S - Y_u \Vert^2 du$, by the supremum of the distances $ \sup_{0\leq t \leq T} \Vert W^S_t - Y_t \Vert^2$, and introducing the parameter $\beta$.

Note that the same geometric distance term $ \sup_{0\leq t \leq T} \Vert W^S_t - Y_t \Vert$ as in \cref{sec:main-result-sup} appears in \cref{thm:pac_bayes_wassertein,thm:disintegrated_heavy_tailed_bound}, but this time to the power $2$.
As mentioned in \cref{sec:main-result-sup}, the term $\sup_{0\leq t \leq T}\Vert W^S_t - Y_t \Vert^2$ can be bounded by methods relying on the fractal properties of the heavy-tailed dynamics. As it is shown in \cref{thm:second_main_result}, it is bounded by a term of magnitude $\nicefrac{C^2_{L,T}}{n}$, where $C_{L,T}$ depends, among other things, on $L$ and $T$. 
By optimizing the value parameter $\lambda$ in the above bounds, we can obtain generalization bounds of order $\mathcal{O}( \sqrt{\nicefrac{\log(1/\zeta)}{n}} + \nicefrac{C_{L,T}}{n} )$, while the bounds in \cref{sec:main-result-sup} are in $\mathcal{O} \left( \nicefrac{C_{L,T}}{\sqrt{n}} \right)$. 
Therefore, the potentially big introduced constants are compensated by a higher order in $n$, therefore improving the bound. Our PAC-Bayesian framework has less dependence on the Lipschitz constant $L$, and even no dependence under \cref{ass:partal_derivatives_concentration}.
Moreover, in \cref{sec:additional_results}, we will present, that under different assumptions, we tighten the results and obtain time-independent bounds.

\section{Conclusion}
\label{sec:conclusion}
In this paper, we derived generalization bounds for understanding the generalization ability of continuous dynamics driven by a stable Lévy process, and depending on the empirical risk.
This highlights the generalization properties of models obtained with a stochastic gradient descent algorithm exhibiting heavy-tailed gradient noises.
The main advantage of our analysis is that it does not depend on a mutual information term that is not fully understandable.
Instead, we considered the expected dynamic (depending on the unknown population risk) and introduced a new geometric term in the results, that we bound using techniques from the heavy-tails and fractal literature. Several points are still to be investigated.
From the theoretical perspective, it would be interesting to elaborate on the ideas presented in \cref{sec:additional_results} to tighten the bounds. Moreover, we would like to extend our methods to discrete dynamics with heavy-tailed noise, and also prove non-uniform bounds (\ie on the last point).

\section*{Acknowledgements}

We thank George Deligiannidis for the valuable feedback and discussions. 
B.D. is partially supported by the European Research Council Starting Grant DYNASTY – 101039676.
P.V. is partially supported by the French government under the management of Agence Nationale de la Recherche as part of the ``Investissements d’avenir'' program, reference ANR-19-P3IA-0001 (PRAIRIE 3IA Institute).

\bibliographystyle{abbrvnat}
\bibliography{main}

\begin{thebibliography}{38}
\providecommand{\natexlab}[1]{#1}
\providecommand{\url}[1]{\texttt{#1}}
\expandafter\ifx\csname urlstyle\endcsname\relax
  \providecommand{\doi}[1]{doi: #1}\else
  \providecommand{\doi}{doi: \begingroup \urlstyle{rm}\Url}\fi

\bibitem[Alquier(2021)]{alquier_user-friendly_2021}
P.~Alquier.
\newblock User-friendly introduction to {PAC-Bayes} bounds.
\newblock \emph{CoRR}, abs/2110.11216, 2021.

\bibitem[Birdal et~al.(2021)Birdal, Lou, Guibas, and {\c S}im{\c
  s}ekli]{birdal_intrinsic_2021}
T.~Birdal, A.~Lou, L.~J. Guibas, and U.~{\c S}im{\c s}ekli.
\newblock Intrinsic dimension, persistent homology and generalization in neural
  networks.
\newblock In \emph{Advances in Neural Information Processing Systems
  (NeurIPS)}, pages 6776--6789, 2021.

\bibitem[Blumenthal and Getoor(1960)]{blumenthal_theorems_1959}
R.~Blumenthal and R.~Getoor.
\newblock Some theorems on stable processes.
\newblock \emph{Transactions of the American Mathematical Society}, 95\penalty0
  (2):\penalty0 263--273, 1960.

\bibitem[B{\"o}ttcher et~al.(2013)B{\"o}ttcher, Schilling, and
  Wang]{bottcher_levy_2013}
B.~B{\"o}ttcher, R.~Schilling, and J.~Wang.
\newblock \emph{Lévy {Matters} {III}: {Lévy}-{Type} {Processes}:
  {Construction}, {Approximation} and {Sample} {Path} {Properties}}, volume
  2099 of \emph{Lecture {Notes} in {Mathematics}}.
\newblock Springer International Publishing, 2013.

\bibitem[Camuto et~al.(2021)Camuto, Deligiannidis, Erdogdu,
  G{\"{u}}rb{\"{u}}zbalaban, {\c S}im{\c s}ekli, and Zhu]{camuto_fractal_2021}
A.~Camuto, G.~Deligiannidis, M.~Erdogdu, M.~G{\"{u}}rb{\"{u}}zbalaban, U.~{\c
  S}im{\c s}ekli, and L.~Zhu.
\newblock Fractal structure and generalization properties of stochastic
  optimization algorithms.
\newblock In \emph{Advances in Neural Information Processing Systems
  (NeurIPS)}, pages 18774--18788, 2021.

\bibitem[Chen et~al.(2023)Chen, Deng, Schilling, and
  Xu]{chen_approximation_2023}
P.~Chen, C.-S. Deng, R.~Schilling, and L.~Xu.
\newblock Approximation of the invariant measure of stable {SDEs} by an
  {Euler}–{Maruyama} scheme.
\newblock \emph{Stochastic Processes and their Applications}, 163:\penalty0
  136--167, 2023.

\bibitem[Dupuis et~al.(2023)Dupuis, Deligiannidis, and {\c S}im{\c
  s}ekli]{dupuis_generalization_2023}
B.~Dupuis, G.~Deligiannidis, and U.~{\c S}im{\c s}ekli.
\newblock Generalization bounds with data-dependent fractal dimensions.
\newblock In \emph{International Conference on Machine Learning (ICML)}, 2023.

\bibitem[Falconer(2014)]{falconer_fractal_2014}
K.~Falconer.
\newblock \emph{Fractal Geometry: Mathematical Foundations and Applications}.
\newblock John Wiley \& Sons, Inc., third edition, 2014.

\bibitem[Germain et~al.(2009)Germain, Lacasse, Laviolette, and
  Marchand]{germain_pac-bayesian_2009}
P.~Germain, A.~Lacasse, F.~Laviolette, and M.~Marchand.
\newblock {PAC-Bayesian} learning of linear classifiers.
\newblock In \emph{International Conference on Machine Learning (ICML)}, volume
  382 of \emph{{ACM} International Conference Proceeding Series}, pages
  353--360. {ACM}, 2009.

\bibitem[G{\"{u}}rb{\"{u}}zbalaban et~al.(2021)G{\"{u}}rb{\"{u}}zbalaban, {\c
  S}im{\c s}ekli, and Zhu]{gurbuzbalaban_heavy-tail_2021}
M.~G{\"{u}}rb{\"{u}}zbalaban, U.~{\c S}im{\c s}ekli, and L.~Zhu.
\newblock The heavy-tail phenomenon in {SGD}.
\newblock In \emph{International Conference on Machine Learning (ICML)}, volume
  139 of \emph{Proceedings of Machine Learning Research}, pages 3964--3975.
  {PMLR}, 2021.

\bibitem[Hellstr{\"{o}}m et~al.(2023)Hellstr{\"{o}}m, Durisi, Guedj, and
  Raginsky]{hellstrom_generalization_2023}
F.~Hellstr{\"{o}}m, G.~Durisi, B.~Guedj, and M.~Raginsky.
\newblock Generalization bounds: {P}erspectives from information theory and
  {PAC-B}ayes.
\newblock \emph{CoRR}, abs/2309.04381, 2023.

\bibitem[Hodgkinson and Mahoney(2021)]{hodgkinson_multiplicative_2021}
L.~Hodgkinson and M.~Mahoney.
\newblock Multiplicative noise and heavy tails in stochastic optimization.
\newblock In \emph{International Conference on Machine Learning (ICML)}, volume
  139 of \emph{Proceedings of Machine Learning Research}, pages 4262--4274.
  {PMLR}, 2021.

\bibitem[Hodgkinson et~al.(2021)Hodgkinson, {\c S}im{\c s}ekli, Khanna, and
  Mahoney]{hodgkinson_generalization_2021}
L.~Hodgkinson, U.~{\c S}im{\c s}ekli, R.~Khanna, and M.~Mahoney.
\newblock Generalization properties of stochastic optimizers via trajectory
  analysis.
\newblock \emph{CoRR}, abs/2108.00781, 2021.

\bibitem[Hodgkinson et~al.(2022)Hodgkinson, {\c S}im{\c s}ekli, Khanna, and
  Mahoney]{hodgkinson_generalization_2022}
L.~Hodgkinson, U.~{\c S}im{\c s}ekli, R.~Khanna, and M.~Mahoney.
\newblock Generalization bounds using lower tail exponents in stochastic
  optimizers.
\newblock In \emph{International Conference on Machine Learning (ICML)}, volume
  162 of \emph{Proceedings of Machine Learning Research}, pages 8774--8795.
  {PMLR}, 2022.

\bibitem[Jain and Pruitt(1968)]{pruitt_correct_1968}
N.~Jain and W.~Pruitt.
\newblock The correct measure function for the graph of a transient stable
  process.
\newblock \emph{Zeitschrift für Wahrscheinlichkeitstheorie und Verwandte
  Gebiete}, 9\penalty0 (2):\penalty0 131--138, 1968.

\bibitem[Knopova et~al.(2015)Knopova, Schilling, and Wang]{knopova_lower_2015}
V.~Knopova, R.~Schilling, and J.~Wang.
\newblock Lower bounds of the {Hausdorff} dimension for the images of {Feller}
  processes.
\newblock \emph{Statistics \& Probability Letters}, 97:\penalty0 222--228,
  2015.

\bibitem[Lugosi and Neu(2022)]{lugosi_generalization_2022}
G.~Lugosi and G.~Neu.
\newblock Generalization bounds via convex analysis.
\newblock In \emph{Conference on Learning Theory (COLT)}, volume 178 of
  \emph{Proceedings of Machine Learning Research}, pages 3524--3546. {PMLR},
  2022.

\bibitem[Mattila(1995)]{mattila_geometry_1995}
P.~Mattila.
\newblock \emph{Geometry of Sets and Measures in {Euclidean} Spaces: {Fractals}
  and Rectifiability}.
\newblock Cambridge University Press, first edition, 1995.

\bibitem[McAllester(1998)]{mcallester_some_1998}
D.~McAllester.
\newblock Some {PAC-B}ayesian theorems.
\newblock In \emph{Conference on Computational Learning Theory (COLT)}, pages
  230--234. {ACM}, 1998.

\bibitem[Mohri et~al.(2012)Mohri, Rostamizadeh, and
  Talwalkar]{mohri_foundations_2012}
M.~Mohri, A.~Rostamizadeh, and A.~Talwalkar.
\newblock \emph{Foundations of Machine Learning}.
\newblock Adaptive computation and machine learning. {MIT} Press, 2012.

\bibitem[Molchanov(2017)]{molchanov_theory_2017}
I.~Molchanov.
\newblock \emph{Theory of Random Sets}, volume~87 of \emph{Probability Theory
  and Stochastic Modelling}.
\newblock Springer London, 2017.

\bibitem[Nguyen et~al.(2019)Nguyen, {\c S}im{\c s}ekli,
  G{\"{u}}rb{\"{u}}zbalaban, and Richard]{nguyen_first_2019}
T.~H. Nguyen, U.~{\c S}im{\c s}ekli, M.~G{\"{u}}rb{\"{u}}zbalaban, and
  G.~Richard.
\newblock First exit time analysis of stochastic gradient descent under
  heavy-tailed gradient noise.
\newblock In \emph{Advances in Neural Information Processing Systems
  (NeurIPS)}, pages 273--283, 2019.

\bibitem[Pruitt(1969)]{pruitt_hausdorff_1969}
W.~Pruitt.
\newblock The {Hausdorff} dimension of the range of a process with stationary
  independent increments.
\newblock \emph{Indiana University Mathematics Journal}, 19\penalty0
  (4):\penalty0 371--378, 1969.

\bibitem[Pruitt and Taylor(1969)]{pruitt_sample_1969}
W.~Pruitt and J.~Taylor.
\newblock Sample path properties of processes with stable components.
\newblock \emph{Zeitschrift für Wahrscheinlichkeitstheorie und Verwandte
  Gebiete}, 12\penalty0 (4):\penalty0 267--289, 1969.

\bibitem[Raj et~al.(2023)Raj, Zhu, G{\"{u}}rb{\"{u}}zbalaban, and {\c S}im{\c
  s}ekli]{raj_algorithmic_2023}
A.~Raj, L.~Zhu, M.~G{\"{u}}rb{\"{u}}zbalaban, and U.~{\c S}im{\c s}ekli.
\newblock Algorithmic stability of heavy-tailed {SGD} with general loss
  functions.
\newblock In \emph{International Conference on Machine Learning (ICML)}, volume
  202 of \emph{Proceedings of Machine Learning Research}, pages 28578--28597.
  {PMLR}, 2023.

\bibitem[Revuz and Yor(1999)]{revuz_continuous_1999}
D.~Revuz and M.~Yor.
\newblock \emph{Continuous Martingales and {Brownian} Motion}, volume 293 of
  \emph{Grundlehren der mathematischen {Wissenschaften}}.
\newblock Springer Berlin Heidelberg, 1999.

\bibitem[Schilling(1998)]{schilling_feller_1998}
R.~Schilling.
\newblock Feller processes generated by pseudo-differential operators: {O}n the
  {H}ausdorff dimension of their sample paths.
\newblock \emph{Journal of Theoretical Probability}, 11\penalty0 (2):\penalty0
  303--330, 1998.

\bibitem[Schilling(2016)]{schilling_introduction_2016}
R.~Schilling.
\newblock An introduction to {L}{\'e}vy and {Feller} processes. {Advanced}
  courses in mathematics - {CRM} {Barcelona} 2014.
\newblock \emph{CoRR}, abs/1603.00251, 2016.

\bibitem[Shawe{-}Taylor and Williamson(1997)]{shawetaylor_pac_1997}
J.~Shawe{-}Taylor and R.~Williamson.
\newblock A {PAC} analysis of a {Bayesian} estimator.
\newblock In \emph{Conference on Computational Learning Theory (COLT)}, pages
  2--9. {ACM}, 1997.

\bibitem[{\c S}im{\c s}ekli et~al.(2019){\c S}im{\c s}ekli, Sagun, and
  G{\"{u}}rb{\"{u}}zbalaban]{simsekli_tail-index_2019}
U.~{\c S}im{\c s}ekli, L.~Sagun, and M.~G{\"{u}}rb{\"{u}}zbalaban.
\newblock A tail-index analysis of stochastic gradient noise in deep neural
  networks.
\newblock In \emph{International Conference on Machine Learning (ICML)},
  volume~97 of \emph{Proceedings of Machine Learning Research}, pages
  5827--5837. {PMLR}, 2019.

\bibitem[{\c S}im{\c s}ekli et~al.(2021){\c S}im{\c s}ekli, Sener,
  Deligiannidis, and Erdogdu]{simsekli_hausdorff_2021}
U.~{\c S}im{\c s}ekli, O.~Sener, G.~Deligiannidis, and M.~Erdogdu.
\newblock Hausdorff dimension, heavy tails, and generalization in neural
  networks.
\newblock \emph{Journal of Statistical Mechanics: Theory and Experiment},
  2021\penalty0 (12), 2021.

\bibitem[Taylor(1967)]{taylor_sample_1967}
J.~Taylor.
\newblock Sample path properties of a transient stable process.
\newblock \emph{Journal of Mathematics and Mechanics}, 16\penalty0
  (11):\penalty0 1229--1246, 1967.

\bibitem[van Erven and Harremo{\"{e}}s(2014)]{van_erven_renyi_2014}
T.~van Erven and P.~Harremo{\"{e}}s.
\newblock R{\'{e}}nyi divergence and {Kullback-Leibler} divergence.
\newblock \emph{{IEEE} Transactions on Information Theory}, 60\penalty0
  (7):\penalty0 3797--3820, 2014.

\bibitem[Vershynin(2018)]{vershynin_high-dimensional_2020}
R.~Vershynin.
\newblock \emph{High-Dimensional Probability: {An} Introduction with
  Applications in Data Science}.
\newblock Cambridge University Press, first edition, 2018.

\bibitem[Viallard et~al.(2023)Viallard, Germain, Habrard, and
  Morvant]{viallard_general_2021}
P.~Viallard, P.~Germain, A.~Habrard, and E.~Morvant.
\newblock A general framework for the practical disintegration of
  {PAC-Bayesian} bounds.
\newblock \emph{Machine Learning}, 2023.

\bibitem[Wang et~al.(2021)Wang, G{\"{u}}rb{\"{u}}zbalaban, Zhu, {\c S}im{\c
  s}ekli, and Erdogdu]{wang_convergence_2021}
H.~Wang, M.~G{\"{u}}rb{\"{u}}zbalaban, L.~Zhu, U.~{\c S}im{\c s}ekli, and
  M.~Erdogdu.
\newblock Convergence rates of stochastic gradient descent under infinite noise
  variance.
\newblock In \emph{Advances in Neural Information Processing Systems
  (NeurIPS)}, pages 18866--18877, 2021.

\bibitem[Xiao(2004)]{xiao_random_2004}
Y.~Xiao.
\newblock Random fractals and {Markov} processes.
\newblock In \emph{Proceedings of {Symposia} in {Pure} {Mathematics}}, volume
  72.2, pages 261--338. American Mathematical Society, 2004.

\bibitem[Xu and Raginsky(2017)]{xu_information-theoretic_2017}
A.~Xu and M.~Raginsky.
\newblock Information-theoretic analysis of generalization capability of
  learning algorithms.
\newblock In \emph{Advances in Neural Information Processing Systems (NIPS)},
  pages 2524--2533, 2017.

\end{thebibliography}

\newpage

\appendix

The appendix is organized as follows:
\begin{itemize}
    \item In \cref{sec:technical_background}, we introduce some mathematical tools necessary for a good understanding of the paper.
    The main goal is to give the definition of Levy and Feller processes, as well as their relation to fractal geometry. This allows us to define precisely the fractal dimensions appearing in our proofs and prove rigorously \cref{thm:heavy_tiled_sdes}.
    We also provide a quick technical background on PAC-Bayesian theory.
    \item \cref{sec:additional_results} presents an additional bound on the distance between empirical and expected trajectories, based on a dissipativity assumption.
    \item In \cref{sec:posponed_proof}, we present all proofs of our results.
\end{itemize}

\section{Technical Background}
\label{sec:technical_background}

In this section, we provide a little technical background in order to present the tools and theorems that will be used in our proofs. 

\subsection{Lévy and Feller Processes}
\label{sec:levy_feller}

In this subsection, we describe some basic notions related to Lévy and Feller processes.
We mention only the tools that are necessary for our results and their proofs, for more technical details and proofs, the reader may look at the tutorial of \citet{schilling_introduction_2016} or the following textbooks \citep{bottcher_levy_2013,xiao_random_2004}.

\textbf{Lévy processes:} A Lévy process $(X_t)_{t\geq0}$ is a stochastic process, in $\mathds{R}^d$, with $X_0 = 0$ almost surely, satisfying the following properties:
\begin{itemize}
    \item \textbf{independent increments:} for any $t_0<t_1<\dots < t_k$, the random variables $X_{t_i} - X_{t_{i+1}}$, for $1 \leq i \leq k$, are independent.
    \item \textbf{Stationary increments:} For any $s < t$ the random variables $X_t - X_s$ and $X_{t-s}$ have the same distribution.
    \item $X$ is \textbf{stochastically continuous}, meaning:
    \begin{align*}
        \forall \epsilon > 0,~\forall s\geq 0, \Pof[]{\Vert X_t - X_s \Vert > \epsilon} \underset{t\to s}{\longrightarrow} 0.
    \end{align*}
\end{itemize}
One can show that such processes have modifications (\ie a process $(\Tilde{X}_t)_{t \geq 0}$ such that, for all $t$, $X_t = \Tilde{X}_t$ almost surely) with almost surely \emph{càdlàg}\footnote{Càdlàg means that the paths are right continuous and admit left limits everywhere.} paths, see \citep[chapter $2$]{schilling_introduction_2016}, it is equivalent to the property of being stochastically continuous.
In this work, we always assume that the considered Levy processes have almost surely càdlàg paths.
Thus, Lévy processes are not continuous but admit \emph{random jumps}.
It should be noted that càdlàg paths are almost surely bounded on compact intervals, which we shall use without mentioning it.

\begin{example}
    Brownian motion is a Lévy process.
\end{example}

Lévy processes are characterized by the following expression of their characteristic function:
\begin{align}
    \label{eq:levy_characteristic}
    \Eof[]{e^{i \xi \cdot X_t}} = e^{-t\psi(\xi)},
\end{align}
where $\psi$ is called the \emph{characteristic exponent}, and is given by the Lévy-Khintchine formula:
\begin{align}
    \psi(\xi) = i b \cdot \xi + \frac{1}{2} \xi^T \Sigma \xi + \int_{\Rd} \left\{  1 - e^{i\xi \cdot x} + i \frac{\xi \cdot x}{1 + \Vert x \Vert^2} \right\} dL(x),
\end{align}
where $b\in \mathds{R}^d$, $\Sigma$ is a symmetric, non-negative definite, $d\times d$ matrix and $L$ is the \emph{Lévy measure} (on Borel sets) satisfying\footnote{Note that there exist several equivalent definitions of Levy measure.}: $\int_{x\neq 0} \frac{|x|^2}{1 + |x|^2} dL(x) < +\infty$.
Intuitively, the first term $i b \cdot \xi$ is a \emph{drift} term, the second ($\nicefrac{1}{2} \xi^T \Sigma \xi$) is a Gaussian term (\ie the characteristic function of a multivariate Gaussian), while the last term encodes information about the jump structure and the (potential) heavy-tailed properties of the process. 

\begin{example}
    \label{example:levy_is_brownian}
    If $b=0$ and $L=0$, the characteristic function reduces to a Gaussian, \ie, the process is a Brownian motion.
\end{example}

The reason why we are interested in the characteristic exponent $\psi$ is because it is deeply related to the Hausdorff dimension of the process, which we use in the proofs of our main results.
In particular, we consider the class of \emph{stable} Lévy processes. 
Those have been extensively studied for their fractal and statistical properties \citep{blumenthal_theorems_1959,pruitt_correct_1968,pruitt_sample_1969,taylor_sample_1967,xiao_random_2004}. 
Stable processes are characterized by their \emph{index} or \emph{tail-exponent}, denoted $\alpha \in (0,2]$, and the following expression of their Lévy measure:
\begin{align}
    d L(x) = \frac{dr}{r^{1+\alpha}} \nu (d \sigma),
\end{align}
with $\nu$ a finite measure on the sphere $\mathds{S}^{d-1}$. One can show \citep{xiao_random_2004}, that, in that case, the characteristic exponent may be expressed as, with a finite measure $M$ on the sphere.
It is defined by
\begin{align}
    \label{eq:stable_exponent}
    \psi (\xi) = i b \cdot \xi + \int_{\mathds{S}^{d-1}} w_\alpha(\xi, y) dM(y) ~~(\alpha=1), 
\end{align}
where $w_\alpha$ is given by \citet[Equation $3.10$]{schilling_introduction_2016}
\begin{align*}
w_\alpha(\xi,y) = \left\{ 
\begin{aligned}
         |\xi\cdot y|^\alpha \left( 1 - i\sgn (\xi\cdot y) \tan \left( \frac{\pi \alpha}{2}\right)\right),\ \ \text{if}\ \alpha\neq 1,\\
        |\xi\cdot y| \left( 1 + \frac{2i}{\pi} \sgn (\xi\cdot y) \log(\xi\cdot y) \right),\ \ \text{if}\ \alpha= 1.
\end{aligned}
\right.
\end{align*}

A stable process is said to be \emph{strictly stable} if $b= 0$ and, if $\alpha=1$, we also require that $\int ydM(y) = 0$.
We recognize in these equations the expressions of characteristic functions of stable distributions.

\begin{remark}
The notion of stability has several equivalent definitions and is related to stable distribution and the self-similarity of the processes, see \citep[Section $3$]{schilling_introduction_2016}.
\end{remark}

\begin{example}
    If the distribution of the process is rotationally invariant and has no drift ($b=0$), then $\psi$ reduces to a simpler form: $\psi(\xi) = C\Vert \xi \Vert^\alpha$, see \citep[Section $2.1$]{xiao_random_2004}. In the context of \cref{eq:empirical_dynamics}, it means that the noise is rotationally invariant. 
\end{example}

\textbf{Feller processes:} Feller processes are an extension of Lévy processes, they are Markov processes characterized by the fact that their Markov-semigroup operators, $P_t f(w)\defeq \mathds{E}^w [f(X_t)]$, satisfy regularity properties making them a ``Feller semigroup''.\footnote{The notation $\mathds{E}^w [\cdot]$ denotes that the process under the expectation was initialized with $w$.}
For more technical details on Feller processes, see \citep{schilling_introduction_2016}, here we just summarize some important properties useful in this work.
First, note that we can assume that the considered Feller processes also have modifications with almost surely càdlàg paths, see \citep[Definition $11.1$]{schilling_introduction_2016}. 
More importantly, a Feller process $(X_t)_{t\geq0}$ is characterized by its \emph{symbol} $\symb$, which is related to the pseudo-differential operator representation of its generator. For Lévy processes, the symbol is just the characteristic exponent. The next theorem shows that we can see the dynamics given by \cref{eq:empirical_dynamics} as Feller processes, see \citep[Theorem $3.8$]{bottcher_levy_2013}.

\begin{theorem}
    \label{thm:sde_existence_solution}
    Let $L_t$ be a $k$-dimensional Lévy process, with exponent $\psi$, and $E: \Rd \longrightarrow \mathds{R}^{d \times k}$ be a bounded and Lipschitz continuous function. Then the Itô SDE:
    \begin{align}
        \label{eq:general_levy_sde}
        d X_t = E(X_{t^-}) dL_t \quad\text{with}\quad X_0 = x,
    \end{align}
    has a unique strong solution which is a Feller process with the symbol $\symb(x, \xi) = \psi(E(x)^T\xi)$.
\end{theorem}

This theorem first ensures that the equations (that are strong Markov processes) considered in this paper have strong solutions. 
As noted by \citet[Theorem $12.11$]{schilling_introduction_2016}, the Lipschitz continuity assumption is only needed to prove that those solutions are Feller processes.
Moreover, the expression of the symbol $\Psi$, given in \cref{thm:sde_existence_solution}, will be used in the proof of \cref{thm:heavy_tiled_sdes}, to bound the Hausdorff dimension of the processes (that is more formally defined in the next subsection). 

\subsection{Fractal Geometry}
\label{sec:fractal_geometry}

Let us very quickly describe the notions of Minkowski and Hausdorff dimension, for more details on fractal geometry, see \citep{falconer_fractal_2014,mattila_geometry_1995} for more details.
One of the main goals of fractal geometry is to define various notions of dimensions, associated with any set in $\Rd$, extending the usual notion of dimension (\eg, for vector spaces or Riemannian sub-manifolds).
In this paper, those dimensions serve as a complexity measure of the sample paths (\ie, trajectories) of Markov processes, which has been extensively studied in the literature \citep{xiao_random_2004}.
In our work, we focus on two of the most popular notions of dimension: the box-counting dimension (\aka Minkowski dimension) and the Hausdorff dimension. Both are based on the notion of $\delta$-cover.

\begin{definition}[Covering of a set]
    \label{def:delta_cover}
    Let $X \subseteq \Rd$ be a set. A $\delta$-cover is a family $(U_i)$ of sets of diameter at most $\delta$, such that $X \subseteq \bigcup_i U_i$.
    If $X$ is bounded, a cover of minimal cardinality is called \emph{minimal}, and the corresponding cardinal is the \emph{covering number}, denoted $N_\delta(X)$.
\end{definition}

Based on the notion of $\delta$-cover of a set, we are now able to define the (lower and upper) Minkowski dimensions that are used to characterize the dynamics in our paper (see \cref{ass:ahlfors_y}).
Moreover, note that the upper Minkowski dimension is the one that appears most naturally in our proofs.

\begin{definition}[Minkowski dimensions]
    Given a bounded set $X \subset \Rd$, we define the lower and upper Minkowski dimensions as:
    \begin{align*}
    \upperbox(X) \defeq \limsup_{\delta \to 0} \frac{\log N_\delta(X)}{\log(1/\delta)}, \quad  \underline{\dim}_B(X) \defeq \liminf_{\delta \to 0} \frac{\log N_\delta(X)}{\log(1/\delta)}.
    \end{align*}
    If their value coincide, the corresponding limit is the Minkowski dimension and denoted $\dim_B(X)$.
\end{definition}

Informally, having a Minkowski dimension $m$ means that, if we reduce $\delta$ by a factor of $\lambda$, then the number of $\delta/\lambda$-balls necessary to cover is multiplied by $\lambda^m$.

\begin{remark}
    \label{rq:closed_open_cover}
    We use the fact that defining the Minkowski dimensions by using closed or open $\delta$-balls does not change the value of the dimension. 
\end{remark}

We end this series of definitions by defining Hausdorff measures and dimensions \citep[Chapter $3$]{falconer_fractal_2014}.

\begin{definition}[Hausdorff measures and dimensions]
    Let $s \geq 0$ and $X \subset \Rd$, its Hausdorff measure is defined as:
    \begin{align*}
    \mathcal{H}^s(X) = \lim_{\delta \to 0} \inf \left\{ \sum_i \text{\normalfont{diam}}(U_i)^s,~(U_i)_i \text{ is a countable $\delta$-cover of $X$} \right\}.
    \end{align*}
    The Hausdorff measure $s \longmapsto \mathcal{H}^s(X)$ is decreasing and only takes values $\infty$ and $0$, except on one ``jump'' point, which is the Hausdorff dimension, \ie, 
    \begin{align*}
    \hausdim(X) \defeq \inf \{ s\geq 0,~\mathcal{H}^s(X) = 0 \}.
    \end{align*}
\end{definition}

In general, one has the following inequality between the two notions of dimension that we are interested in in this paper, for a bounded set $X \subset \Rd$:
\begin{align}
    \label{eq:ineq_minko_hausdorff}
    \hausdim(X) \leq \upperbox(X).
\end{align}

The following proposition gives a sufficient condition for Hausdorff and upper box-counting dimensions to be equal, see \citep[Theorem $5.7$]{mattila_geometry_1995}:

\begin{proposition}
    \label{prop:alhfors}
    Let $\wcal \subset \Rd$ be bounded and $\pi$ a probability measure on $\wcal$. If there exists, $a,b,r_0,s>0$ such that, for all $x{\in}A$ and $0<r\leq r_0$, we have $ar^s\leq\pi(B(x{,}r))\leq br^s$, then:
    \begin{align*}
    \upperbox(\wcal) = \hausdim(\wcal).
    \end{align*}
\end{proposition}

Therefore, \cref{prop:alhfors} gives a necessary condition so that \cref{ass:ahlfors_y} could be satisfied.

It is known that the Hausdorff dimension of an $\alpha$-\emph{strictly} stable Lévy process is precisely the tail-index $\alpha$ \citep{taylor_sample_1967,pruitt_sample_1969,xiao_random_2004}.
Let us now describe quickly the fractal properties of Feller processes, which we will express in terms of their symbol $\symb$. The following theorem gives a general upper bound of the Hausdorff dimension, see \citep[Theorem $5.15$]{bottcher_levy_2013} and \citep[Theorem $1.2$]{knopova_lower_2015}.

\begin{theorem}
    \label{thm:feller_process_beta_infty_index}
    Let $(X_t)$ be a Feller process with symbol $\symb(x,\xi)$ and infinitesimal generator $A$ (see \citep[Chapter $5$]{schilling_introduction_2016}). Assume that:
    \begin{enumerate}[label={\it (\roman*)}]
    \item $\mathcal{C}_c^\infty \subseteq \mathcal{D}(A)$, where $\mathcal{D}(A)$ is the domain of the generator $A$, where $\mathcal{C}_c^\infty$ holds for smooth functions with compact support (in $\Rd$).
    \item $\symb(\cdot, 0) = 0$ and $|\symb(x,\xi)| \leq c (1 + |\xi|^2)$, for some constant $c$,
    \end{enumerate}
    then
    \begin{align*}
    \hausdim(X([0,T])) \leq \beta_\infty \defeq \inf \left\{ \lambda > 0, ~\lim_{|\xi| \to \infty} \frac{\sup_{|\eta|\leq |\xi|} \sup_{x \in\Rd} |\symb(x,\eta)|  }{|\xi|^\lambda} = 0 \right\},
    \end{align*}
where $\beta_\infty$ is the \emph{generalized Blumenthal–Getoor index at infinity}.
\end{theorem}

As we will see in the proof of \cref{thm:heavy_tiled_sdes}, this theorem is only able to show that $\hausdim(\ycal) = \alpha$ for $\alpha \in [1,2]$.
To handle the case $\alpha < 1$, we can use the following theorem, which essentially, in our case, will require a lot of regularity from $w\longmapsto \nabla_w \ell(w,z)$, see \citep[Theorem $4$]{schilling_feller_1998}.
Note that this theorem is also the one used in \citep{simsekli_hausdorff_2021} to relate the tail index to the Hausdorff dimension of the process, however, it does not seem to be general enough in our particular case, because of the above-mentioned regularity.
The following theorem is a \emph{slightly weaker} formulation of \citep[Theorem $4$]{schilling_feller_1998}.

\begin{theorem}[Schilling's theorem]
    \label{thm:schiling_theorem}
    Let $(X_t)_{t\geq 0}$ be a Feller process with symbol $\symb(x,\xi)$ such that it can be decomposed as $\symb(x,\xi) = q_1(\xi) + q_2(x, \xi)$, and the following conditions hold:
    \begin{enumerate}[label={\it (\roman*)}]
        \item $\symb(x,0) = 0$ and $q_2(\cdot,\xi) \in \mathcal{C}^{d+1}(\Rd)$,
        \item $1 + q_1(\xi) \leq C (1 + \normof{\xi}^2)$, where $C>0$ is some constant.
        \item For $\boldsymbol{i} \in \mathds{N}^d$, if $|\boldsymbol{i}| \leq d+1$, then $|\partial_x^{\boldsymbol{i}}q_2(x,\xi)| \leq \Phi_{\boldsymbol{i}}(x) (1 + \normof{\xi}^2)$, with $\Phi_0 \in L^\infty(\Rd)$ and $\Phi_{\boldsymbol{i}} \in L^1(\Rd)$ for $\boldsymbol{i} \neq 0$,
    \end{enumerate}
    then
    \begin{align*}
    \hausdim(X([0,T])) \leq \betatail \defeq \inf \left\{ \lambda \geq 0,~\lim_{\normof{\xi} \to \infty} \frac{|q_1(\xi)|}{\normof{\xi}^\lambda} = 0 \right\},
    \end{align*}
    where $\beta_{\text{tail}}$ is called the Blumenthal–Getoor index.
\end{theorem}

\subsection{PAC-Bayesian Bounds}
\label{sec:pac_bayes_background}

In this section, we introduce two PAC-Bayesian bounds in the literature that are used in the proofs.
We first present a PAC-Bayesian bound originally proven by \citet[Theorem 2.1]{germain_pac-bayesian_2009}. 
We will see further that the bound contains a function $\Phi$ that must be defined in order to obtain a bound with the (usual) generalization gap.
In the following, we consider a family $(\rho_S)_{S \in \zcal^n}$ (for all $n$), called the posteriors and satisfying the following assumption, making it a Markov kernel:

\begin{enumerate}[label={\it (\roman*)}]
    \item For all $S$, $\rho_S$ is a probability distribution on $\Rd$
    \item For all $B \in \mathcal{B}(\Rd)$, the mapping $S \longmapsto \rho_S(B)$ is measurable, where $\mathcal{B}(\Rd)$ denotes the Borel $\sigma$-algebra of $\Rd$.
\end{enumerate}

\begin{theorem}[General PAC-Bayesian bound]\label{thm:kl_pac_bayes__bound}
Given a function $\Phi: \Rd \times \Zcal^n \to \R$, with probability at least $1{-}\zeta$ over $S \sim \mu_z^{\otimes n}$ we have
\begin{align*}
\Eof[w \sim\rho_{S}]{\Phi(w, S)} \le \log(1/\zeta) + \klb{\rho_{S}}{\pi} + \log \mathds{E}_S \Eof[w \sim \pi]{e^{\Phi(w, S)}},
\end{align*}
where 
\begin{align*}
\klb{\rho_{S}}{\pi} = \int \ln\left(\frac{d\rho_{S}}{d\pi}(w)\right) d\rho_{S}(w)
\end{align*}
is the KL divergence between the posterior distribution $\rho_{S}$ and the prior distribution $\pi$, and $\frac{d\rho_{S}}{d\pi}(w)$ is the Radon-Nikodym derivative, defined when $\rho_{S} \ll \pi$.
\end{theorem}

Put into words, the bound is on the expectation of $\Phi$ over $\rho_{S}$ and depends mainly on the KL divergence and the term $\log \mathds{E}_S \Eof[\pi]{e^{\Phi(w, S)}}$ that is further upper-bounded when instantiating $\Phi$.
Note that the theorem of \citet{germain_pac-bayesian_2009}  is originally presented with a function representing a gap between the population risk and the empirical risk, however, their theorem holds with such a function $\Phi: \Rd \times \Zcal^n \to \R$.

When we consider a bound holding with high probability over $S\sim\mu_z^{\otimes n}$ and $w\sim\rho_{S}$, we use the bound of \citet{viallard_general_2021} (that is called a disintegrated PAC-Bayesian bound).

\begin{theorem}[Disintegrated PAC-Bayesian bound]
\label{thm:disintegrated_pac_bayes__bound}
Given a function $\Phi: \Rd \times \Zcal^n \to \R$, we have, with probability at least $1 - \zeta$ over $S\sim \mu_z^{\otimes n}$ and $w \sim \rho_S$:
\begin{align*}
 \frac{\beta}{\beta{-}1}\Phi(w, S) \le \frac{2\beta-1}{\beta-1}\log(2/\zeta) + \renyi[\beta]{\rho_{S}}{\pi} + \log \mathds{E}_S \Eof[w \sim \pi]{e^{\frac{\beta}{\beta-1}\Phi(w, S)}},
\end{align*}
where 
\begin{align*}
\renyi[\beta]{\rho_{S}}{\pi} = \frac{1}{\beta-1} \log\left(\int\left(\frac{d\rho_{S}}{d\pi}(w)\right)^\beta d\pi(w)\right)
\end{align*}
is the Rényi divergence between $\rho_{S}$ and $\pi$.
\end{theorem}
Similarly as for \cref{thm:kl_pac_bayes__bound}, the function $\Phi$ must be defined and the term $\log \mathds{E}_S \Eof[w \sim \pi]{e^{\frac{\beta}{\beta-1}\Phi(w, S)}}$ must be upper-bounded to obtain a bound with the generalization error; this is actually done in the proofs.

\subsection{Egoroff's theorem}

In our proofs, we will use the following theorem, to get uniform convergence of the limits defining the Minkowski dimensions. 
This has been used in other works \citep{simsekli_hausdorff_2021,camuto_fractal_2021,dupuis_generalization_2023}.

\begin{theorem}[Egoroff's theorem]
    \label{thm:Egoroff_theorem}
    Let $(\Omega, \mathcal{F}, \prob)$ be a probability space and $f, (f_n)_{n\in \mathds{N}}$ be measurable functions. Assume that, for almost all $x \in \Omega$, we have $f_n(x) \underset{n \to \infty}{\longrightarrow} f(x)$. Then, for any $\epsilon > 0$ there exists $\Omega_\epsilon \in \mathcal{F}$ such that $\Pof[]{\Omega_\epsilon} \geq 1 - \epsilon$ and on which the convergence of $f_n$ to $f$ is uniform.
\end{theorem}

\section{Additional Results}
\label{sec:additional_results}

The goal of this section is to show that, under an additional reasonable assumption, we may render our PAC-Bayesian bounds time-independent.
The following assumption is similar to the so-called ``dissipativity'' assumption, commonly used in the stochastic calculus literature.
This special form is more adapted to the simultaneous consideration of two dynamics and has recently been made by \citet{chen_approximation_2023} and \citet{raj_algorithmic_2023}.

\begin{assumption}[Co-dissipativity]
    \label{ass:co_dissipativity}
    There exists $m > 0$ and $K\geq 0$ such that:
    \begin{align*}
    \langle \nabla\ell(w',z) - \nabla \ell(w,z), w - w' \rangle \leq K - m \Vert w - w' \Vert^2,
    \end{align*}
    where all gradients are with respect to the $w$ variable.
\end{assumption}

This assumption will be used to ensure that the difference between the empirical and expected dynamics does not become too large.
Under this assumption, we can tighten the bound on the distance between the empirical and the expected dynamics.

\begin{restatable}[Bound under co-dissipativity]{theorem}{thmCoDissipativity}
    \label{thm:co_dissipativity_bound_on_sup}
    We make \cref{ass:sub_Gaussian,ass:co_dissipativity} and with a loss $\ell$ that is $M$-smooth in $w$ (but not Lipschitz). We further assume that there exists $w_0 \in \Rd$ such that $\nabla \ell (w_0, z) \in L^1(\mu_z)$.
    Then, for $n$ large enough, with probability at least $1 - \zeta$ over $\mu_u \otimes \mu_z^{\otimes n}$, we have, almost surely with respect to $(S,U)$:
    \begin{align*}
        \sup_{0 \leq t \leq T}  \Vert W_t^S - Y_t \Vert^2 \leq \frac{2K}{m} + \frac{4}{m^2} \bigg\{ \frac{2}{n} + \frac{\Sigma^2}{n} \left( 2\dimvalue \log(M\sqrt{n}) + \log(2d/\zeta) \right) \bigg\}.
    \end{align*}
\end{restatable}

\begin{remark}
    \label{rq:not_lipschitz_not_feller}
    If, as it is the case in the above theorem, we do not make the Lipschitz assumption on the loss $\ell$, we cannot ensure anymore that $W^S$ and $Y$ are Feller processes, see \citep[Theorem $12.11$]{schilling_introduction_2016}.
    However, we can still argue that the process has a càdlàg modification, as a consequence of martingale regularization theorems \citep[Section $2.2$]{revuz_continuous_1999}, and hence we can show that the process has almost surely bounded paths.
    This makes that the upper box-counting dimensions, $\upperbox(W^S([0,T]))$ and $\upperbox(Y([0,T]))$ are well defined.
    However, we may not be able to formally relate it to $\alpha$. In the above theorem, the notation $\gamma$ therefore refers to the upper box-counting dimension.
\end{remark}

\section{Postponed Proofs}
\label{sec:posponed_proof}

\subsection{Technical Lemmas}
\label{sec:technical_lemmas}

In this subsection, we quickly prove a few technical results, especially measure theoretic ones.
The goal is to ensure solid theoretical foundations for our work.
Some of the results of this section are used without further notice throughout the proofs.

\begin{lemma}[Almost surely càdlàg paths]
$Y$ and $W^S$ are almost surely càdlàg \wrt $\mu_z^{\otimes n} \otimes \mu_u$.
\end{lemma}
\begin{proof}
    As mentioned in \cref{rq:well-posedness}, we have almost surely, under $\mu_z^{\otimes n} \otimes \mu_u$:
    \begin{align*}
    W_t^S = W_0 - \int_0^t \nabla \er(W_u^S) du + \levy.
    \end{align*}
    By \cref{ass:smooth_lipschitz}, the gradient $\nabla \er$ is bounded, so that the integral above is a continuous function of $t$, as $\levy$ is an almost surely càdlàg version, it ensures that $W^S$ is almost surely càdlàg with respect to $\mu_z^{\otimes n} \otimes \mu_u$. We can do the same reasoning for $Y$.
\end{proof}

\begin{remark}
    As mentioned in \citep[Section $11$]{schilling_introduction_2016}, we could have argued that $W^S$ and $Y$ are Feller processes, and therefore we may take càdlàg versions of them.
    However, for measure-theoretic reasons, it may be safer to directly use the Lipschitz assumption as we did.
\end{remark}

\begin{remark}
    If we do not make the Lipschitz assumption, we could still prove that the processes $W^S$ and $Y$ have a càdlàg modification, by using the fact that they are (local) semi-martingales, see \citep[Theorem $12.11$]{schilling_introduction_2016}.
\end{remark}

We define the difference process as (the dependence on $U$ is omitted):
\begin{align}
    \label{eq:difference_process}
    V_t^S = W_t^S - Y_t
\end{align}

One important result is the following:

\begin{lemma}[Joint continuity of the difference process]
    The map $t \longmapsto V_t^S$ is almost surely continuous under $\mu_z^{\otimes n} \otimes \mu_u$.
\end{lemma}

\begin{proof}
    We have, according to \cref{rq:well-posedness}, almost surely under $\mu_z^{\otimes n} \otimes \mu_u$ that:
    \begin{align*}
    V_t^S = \int_0^t \left( -\nabla \er(W^S_u) + \nabla \risk(Y_u) \right) du
    \end{align*}
    Therefore $t \longmapsto V_t^S$ is almost surely continuous.
    It may be seen either using the boundedness of the gradient, as in the previous proof, or by the fact that càdlàg paths are bounded on compact intervals.
    Note that the fact that both SDEs have the same initialization is crucial here.
\end{proof}

\begin{lemma}
    Under assumptions \ref{ass:smooth_lipschitz}, the processes $W^S$ and $Y$ are Feller processes.
\end{lemma}

\begin{proof}
    This is a direct consequence of \cref{ass:smooth_lipschitz} and \cref{thm:sde_existence_solution}.
\end{proof}

Let's continue with a remark regarding covering arguments made in this paper.
We will use minimal $\delta$-covers of the trajectory $\ycalu$ of $Y$. 
By using arguments similar to \citet{dupuis_generalization_2023}, we show that we can get measurable covering numbers that yield the correct Minkowski dimensions.
This justifies the validity of subsequent covering arguments.

\begin{lemma}
    Let us fix some $\delta > 0$ and $T>0$, then the mapping $U \longmapsto |N_\delta(\overline{\ycalu})|$ is measurable, if open balls are used for the definition of minimal coverings\footnote{According to \cref{rq:closed_open_cover}, this does not change the value of the dimensions}.
\end{lemma}

\begin{proof}
    The proof works by constructing a kind of ``Castaing representation'' \citep[Definition $1.3.6$]{molchanov_theory_2017}. For any $q \in \mathds{Q}_+$, we set $\xi_q(U) \defeq Y_q(U)$.
    By the almost surely càdlàg property of $Y$, it is clear that $\overline{\{ \xi_q \}} = \ycal$, almost surely.
    Moreover, with $\mathds{Q}_T \defeq \mathds{Q}\cap[0,T]$:
    \begin{align*}
    |N_\delta(\overline{\ycalu})| > N = \bigcap_{F \in \mathds{Q}_T^N} \bigcup_{q \in \mathds{Q}_T} \bigcap_{x \in F} \{ |Y_q(U) - Y_x(U)| \geq \delta \},
    \end{align*}
    which, together with the completeness of the space $(\Omega, \mathcal{T}, \mu_u)$, implies the desired measurability.
\end{proof}

To end this technical section, we mention the measurability of the mapping $(S,U) \longmapsto \wgen$, which is one of the main interests of our work.
From the almost surely càdlàg property of the process, we have, almost surely:
\begin{align*}
\wgen = \sup_{t\in \mathds{Q}\cap[0,T]} \left( \risk(W^S_t(U)) - \er(W^S_t(U)) \right),
\end{align*}
which, by measurability of $(t,S,U) \longmapsto W^S_t(U))$, implies the desired measurability, if the space $\zcal^n \times \Omega$ is complete.
We can assume it is the case, in the worst case, considering its completion.

\subsection[Proof of Th. 1]{Proof of \cref{thm:heavy_tiled_sdes}}

In the rest of the paper, we will use the notation $\partial_j$ as a shortcut for the partial derivative $\partial/\partial w_j$, all considered partial derivatives are implicit with respect to $w$, even if it is not mentioned.
Moreover, given a multi-index $\multiindex \in \mathds{N}^d$, the notation $\partial^\multiindex_w$ is a shortcut for $\frac{\partial}{\partial^{i_1}w_1 \dots \partial^{i_d}w_d}$.
We also denote $|\multiindex| = i_1 + \dots + i_d$.

\thmFractalProperties*

Before proving this theorem, let us give its more formal version.

\begin{theorem}[Fractal properties of heavy-tailed dynamics]
    \label{thm:heavy_tiled_sdes_formal}
    Under \cref{ass:sub_Gaussian,ass:smooth_lipschitz}, \cref{eq:empirical_dynamics,eq:expected_dynamics}, if one of the following condition is true:
    \begin{enumerate}[label={\it (\roman*)}]
        \item $\alpha \in [1,2]$, or
        \item $L_t^\alpha$ is strictly stable and, for every $z\in\zcal$ and every multi-index $\multiindex \in \mathds{N}^d$, with $1 \leq |\multiindex| \leq d+1$, the function $w \longmapsto \normof{\partial_w^{\multiindex} \nabla \ell(w,z)}$ is in $L^1(\Rd)$,
    \end{enumerate}
    then $\hausdim(\ycalu), \hausdim(\wcalsu) \leq  \alpha$, hence, under \cref{ass:ahlfors_y}, $\gamma \leq \alpha$.
\end{theorem}

\begin{proof}
    In this proof, $F$ will denote a function that can be both $\risk$ or $\er$, without changing any of the results or computations.
    In the same way, the Feller process solution of the associated SDE (existence ensured by \cref{thm:sde_existence_solution}) will be denoted by $(X_t)$, it is therefore either $Y_t$ or $W^S_t$.

    Let us denote by $\psi_\alpha$ the characteristic exponent\footnote{Like in \citep{xiao_random_2004}, we implicitly assume that all stable distributions are non-degenerate} of $L_t^\alpha$.
    The Itô SDEs \cref{eq:empirical_dynamics,eq:expected_dynamics} may be written in the form of \cref{eq:general_levy_sde}, with $L_t \defeq (L_t^\alpha, t)$ and $E(x) = (I_d, -\nabla F(x)) \in \R^{d \times (d+1)}$.
    From \cref{ass:smooth_lipschitz}, $\nabla F$ is Lipschitz and bounded, hence, by \cref{thm:sde_existence_solution}, the symbol of the solution of the equation, $X_t$, is:
    \begin{align*}
        \symb(x,\xi) = \psi(\xi) - i \nabla F(x) \cdot \xi.
    \end{align*}

    \textbf{Case $1$:} when $\alpha \in [1,2]$. 
    We look at the generalized Blumenthal–Getoor index of \cref{thm:feller_process_beta_infty_index}. Let $\lambda > \alpha$, we have, by the Lipschitz assumption (\cref{ass:smooth_lipschitz}):
    \begin{align*}
    \sup_{|\eta|\leq |\xi|} \sup_{x \in\Rd} |\symb(x,\eta)| \leq \sup_{|\eta|\leq |\xi|} \left(|\psi(\eta)| + L\normof{\eta} \right)
    \end{align*}
    From the explicit expressions of the characteristic exponent of stable processes, given in \citep[Section $2.1$]{xiao_random_2004} and \citep{pruitt_sample_1969}, we have 
    \begin{align*}
    |\psi(\xi)| \leq C_\alpha \normof{\xi}(1 + \log\normof{\xi}) + \mu_0 \normof{\xi}
    \end{align*}
    (the log term handles the case of a Cauchy process), where $C$ and $\mu_0$ are constants. From this we deduce that $\beta_\infty \leq \alpha$, hence, by \cref{thm:feller_process_beta_infty_index}, we have:
    \begin{align*}
    \hausdim(X([0,T])) \leq \alpha.
    \end{align*}
    \textbf{Case $2$:} In that case, even if the process is strictly stable, the drift term in the symbol $\symb$, coming from the SDE, makes that the previous reasoning proves that $\hausdim(L_t^{\alpha}([0,T])) \leq 1$, if $\alpha < 1$.
    To relate the Hausdorff dimension $\hausdim(X([0,T]))$ to $\alpha$ in that case, we make the assumptions of the second part of the theorem.
    As it is now assumed that $L_t^\alpha$ is now strictly stable, it is known that we have:
    \begin{align*}
    \hausdim(L_t^{\alpha}([0,T])) \leq \betatail = \alpha;
    \end{align*}
    see \citep[Theorem $5.12$]{bottcher_levy_2013} or \citep{pruitt_hausdorff_1969,xiao_random_2004}.
    Thanks to the smoothness assumptions on $\nabla_w \ell$, \cref{thm:schiling_theorem}, with $q_1 = \psi$, directly implies that:
    \begin{align*}
    \hausdim(X([0,T])) \leq \alpha.
    \end{align*}
\end{proof}

\subsection[Proof of Th. 2]{Proof of \cref{thm:lipschitz_master_theorem}}

As mentioned earlier, we will prove a slightly stronger version of \cref{thm:lipschitz_master_theorem}, featuring the Hausdorff distance between both trajectories.  
Let us recall the definition of Hausdorff distance.

\begin{definition}[Hausdorff distance]
    \label{def:hausdorff_distance}
    Let $A,B \subset \Rd$ be two non-empty bounded sets, the Hausdorff distance between $A$ and $B$ is defined by:
    \begin{align*}
        d_H(A,B) = \max \left\{ \sup_{a\in A} d(a,B),~\sup_{b\in B} d(b,A) \right\},
    \end{align*}
    where $d(a,B)$ denotes the (Euclidean) distance between the point $a$ and the set $B$ \ie
    \begin{align*}
    d(a,B) = \inf_{b \in B} \normof{a - b}.
    \end{align*}
\end{definition}

We are now ready to prove \cref{thm:lipschitz_master_theorem}.

\theoremhausdorff*

\begin{proof}
    First note that we have:
    \begin{align*}
        \sup_{w \in \wcalsu} d(w, \ycalu) =\sup_{0\leq t \leq T} d(W^S_t, \ycalu) \leq \sup_{0\leq t \leq T} \Vert W^S_t -  Y_t \Vert = \sup_{0\leq t \leq T} \Vert W^S_t - Y_t \Vert ,
    \end{align*}
    and idem for $d(\wcalsu, y)$, thus we have
    \begin{align*}
    d_H (\wcalsu, \ycalu) \leq  \sup_{0\leq t \leq T} \Vert W^S_t - Y_t\Vert .
    \end{align*}
    Therefore, it is enough to prove the bound for the Hausdorff distance, instead of $\sup_{0\leq t \leq T} \Vert W^S_t - Y_t\Vert $.
    Note that the Hausdorff distance $d_H(\wcalsu, \mathcal{Y}_U)$ is almost surely finite because of the càdlàg property of the paths (or their modification).
    Indeed, càdlàg paths are bounded \citep[Section $9$]{schilling_introduction_2016}.
    
    Let us fix some $\eta > 0$ and $S \in \zcal^n$, by definition of the Hausdorff distance, for all $w \in \wcalsu$, there exists $y \in \ycalu$ such that $\Vert w - y \Vert \leq d_H(\wcalsu, \mathcal{Y}_U) + \eta$.
    We also have, by Lipschitz regularity:
\begin{align*}
     \risk(w) - \er(w) &= \risk(w) - \er(w) + \er(y) - \er(y) + \risk(y) - \risk(y)\\
     &\leq |\risk(w) - \risk(y)| + |\er(y) - \er(w)| +  \risk(y) - \er(y)\\
     &= 2  L\Vert w - y \Vert + \risk(y) - \er(y).
\end{align*}
    Therefore, we have
    \begin{align*}
        \sup_{w \in \mathcal{W}_{S,U}} \left( \risk(w) - \er(w) \right) \leq 2L d_H(\wcal_S, \mathcal{Y}) + 2L\eta + \sup_{w \in \ycalu} \left( \risk(w) - \er(w) \right).
    \end{align*}
From there, we bound the term $\sup_{w \in \ycalu} \left( \risk(w) - \er(w) \right)$ using a reasoning similar to the proof of \citep[Theorem $1$]{simsekli_hausdorff_2021}.
We fix some decreasing sequence $(\delta_n)$ such that $\delta_n > 0$. Let $w_1,\dots,w_{N_{\delta_n}}$ be the centers of a minimal cover of $\ycal_U$ with balls of radius at most $\delta_n$ (note that the covering number is random because of its dependence on $U$, when necessary we will denote it $N_{\delta_n}(U)$).
By the same calculation as above, we have:
    \begin{align*}
         \sup_{w \in \ycal_U} \left( \risk(w) - \er(w) \right) \leq 2L\delta_n +  \max_{1 \leq i \leq N_{\delta_n}} \left( \risk(w_i) - \er(w_i) \right).
    \end{align*}
    By Hoeffding's inequality, along with a union bound, we get, for any $\epsilon > 0$:
    \begin{align*}
   \mathds{P}_S \left( \max_{1 \leq i \leq N_{\delta_n}} \left( \risk(w_i) - \er(w_i) \right) \geq \epsilon \right) \leq N_{\delta_n} (U) \exp \left\{ -\frac{n\epsilon^2}{2\sigma^2} \right\}.
    \end{align*}
    Let $\zeta \in (0,1)$, we set:
    \begin{align*}
    \epsilon(U) \defeq \sqrt{2} \sigma \sqrt{\frac{\log(N_{\delta_n} (U)) + \log(1/\zeta)}{n}},
    \end{align*}
    and $\delta_n = 1/(L\sqrt{n})$. By definition of the upper Minkowski dimension, we know that:
    \begin{align*}
    \lim_{k \to \infty}\sup_{n \geq k} \frac{\log(N_{\delta_n} (U))}{\log(1/\delta_n)} = \upperbox(\ycal_U)
    \end{align*}
    By Egoroff's theorem (see \cref{thm:Egoroff_theorem}), there exists $\Omega_\zeta \in \mathcal{T}$, such that $\mu_u(\Omega_\zeta) \geq 1 - \zeta$ and on which the above convergence is uniform in $U$.
    Moreover, thanks to \cref{ass:ahlfors_y}, we have $\upperbox(\ycal_U) = \dimvalue > 0$. 
    From all this we deduce that there exists $N_\zeta$, depending only on $\zeta$, such that, for all $n \geq N_\zeta$, we have that the limit is smaller than $2\gamma$, and therefore:
    \begin{align*}
    \log(N_{\delta_n} (U)) \leq 2\log(L\sqrt{n}) \dimvalue.
    \end{align*}
    Which implies, for all $n \geq N_\zeta$:
    \begin{align}
        \sup_{w \in \mathcal{W}_{S,U}} \left( \risk(w) - \er(w) \right) \leq 2L d_H(\wcal_S, \mathcal{Y}) + 2L\eta + \frac{2}{\sqrt{n}} + \sqrt{2} \sigma \sqrt{\frac{2\log(L\sqrt{n}) \dimvalue + \log(1/\zeta)}{n}}.
        \label{eq:bound-hauss-eta}
    \end{align}

    As this is true for all $\eta$ without changing any of the quantity appearing in \cref{eq:bound-hauss-eta}, we can take $\eta \to 0$ and rearrange the terms to get that, for all $n \geq N_\zeta$:

     \begin{align*}
        \sup_{w \in \mathcal{W}_{S,U}} \left( \risk(w) - \er(w) \right) \leq 2L d_H(\wcal_S, \mathcal{Y})   + \frac{2}{\sqrt{n}}  + 2\sigma \sqrt{\frac{2\log(L\sqrt{n}) \dimvalue + \log(1/\zeta)}{2n}}.
    \end{align*}
    By a union bound, the above event occurs with probability $1 - 2\zeta$ (it is the combination of the probabilities coming from Hoeffding's inequality and Egoroff's theorem).
\end{proof}

\subsection[Proof of Th. 3]{Proof of \cref{thm:second_main_result}}

Before proving \cref{thm:second_main_result}, we present a lemma to bound the following quantity, which is the worst-case deviation of the empirical gradient from its expectation over the trajectory $\ycalu$.
It is defined by

\begin{align}
    \label{eq:gnabla}
    G_\nabla \defeq \gnabla.
\end{align}

\begin{remark}
    Note that thanks to the assumptions, we have boundedness of the gradients $\nabla_w \ell$.
    Therefore, by the dominated convergence theorem, we will be able to invert the expectation and the partial derivative, which we will do repeatedly in our proofs.
\end{remark}

\cref{lemma:bounding_gnabla} will be used for the proof of \cref{thm:second_main_result}.

\begin{lemma}
    \label{lemma:bounding_gnabla}
    For $n$ big enough, with probability at least $1 - \zeta$ over $\mu_u \otimes \mu_z^{\otimes n}$, we have:
    \begin{align*}
    \gnabla \leq \frac{2}{\sqrt{n}} + L \sqrt{\frac{2}{n}} + 2L \sqrt{\frac{2\dimvalue \log(M\sqrt{n}) + \log(1/\zeta)}{2n}}.
    \end{align*}
    If, in addition, we make \cref{ass:partal_derivatives_concentration}, then the result becomes:
    \begin{align*}
    \gnabla \leq \frac{2}{\sqrt{n}} +  2 \Sigma \sqrt{\frac{2\dimvalue \log(M\sqrt{n}) + \log(2d/\zeta)}{2n}}.
    \end{align*}
\end{lemma}

In the proofs and the rest of the document, the notation $\partial_j$ will be used as a shortcut for $\frac{\partial}{\partial w_j}$.

\begin{proof}
We introduce the notation, for $S = (z_,\dots,z_n)$:
\begin{align*}
\overline{\partial_j}(w,z_i) \defeq \partial_j \ell(w,z_i) - \Eof[z]{\partial_j(w,z)}.
\end{align*}
As in the proof of \cref{thm:lipschitz_master_theorem}, let us introduce a sequence $(\delta_n)$, positive, decreasing and converging to $0$, as well as $w_1,\dots,w_{N_{\delta_n}}$ the centers of a minimal cover of $\ycalu$ with balls of radius at most $\delta_n$.
As before, this covering number still depends on $U$.We will denote
    \begin{align*}
    D_S(w) \defeq \Vert \nabla \er(w) - \nabla \risk(w) \Vert.
    \end{align*}
    We have, by Jensen's inequality, independence, and the Lipschitz loss assumption:
    \begin{align*}
        \mathds{E}_S [D_S] &\leq  \sqrt{\E[\Vert \nabla \er(w) - \nabla \risk(w) \Vert^2]} \\
        &= \frac{1}{n} \left\{ \sum_{j=1}^d \Eof{\sum_{1\leq i,k \leq n } \partialmean(w,z_i) \partialmean(w,z_k) }  \right\}^{\frac{1}{2}}\\
        &= \frac{1}{n} \left\{ \sum_{j=1}^d \Eof{\sum_{i=1}^n \partialmean(w,z_i)^2}  \right\}^{\frac{1}{2}}\\
        &= \frac{1}{n} \left\{ \Eof[]{\sum_{i=1}^n \Vert\nabla \ell(w,z_i) - \nabla \risk(w) \Vert^2}\right\}^{\frac{1}{2}}\\
        &\leq L \sqrt{\frac{2}{n}}.
    \end{align*}
    Moreover, if $S^i$ denotes $S\in\zcal^n$, but with $i$-th element $z_i$ replaced by some $z_i' \in \zcal$, we have, by the inverted triangle inequality and the Lipschitz assumption on the loss:
    \begin{align*}
        |D_S(w) - D_{S^i}(w)| \leq \Vert \nabla \er(w) - \nabla \widehat{\mathcal{R}_{S^i}}(w) \Vert \leq \frac{2L}{n}.
    \end{align*}
    Therefore, by McDiarmid inequality, for any $w$, we have:
    \begin{align*}
        \mathds{P}_S \left( D_S(w) - \Eof[]{D_S(w)} \geq \epsilon \right) \leq \exp \left\{ -\frac{n \epsilon^2}{2 L^2} \right\}
    \end{align*}
    The rest of this part of the proof is similar to what we did in the proof of \cref{thm:lipschitz_master_theorem}.
    We apply a covering argument similar to the proof of \cref{thm:lipschitz_master_theorem}, but this time using the $M$-smoothness of $\ell$. Let $w \in \ycalu$, and $(w_1, \dots, w_{N_{\delta_n}})$ be the centers of a $\delta_n$-cover of $\ycalu$, we have that there exists $i$ such that $\normof{w - w_i} \leq \delta_n$, therefore, by smoothness:
    \begin{align*}
        \Vert \nabla \er (w) - \nabla \risk(w) \Vert \leq 2M\delta_n + D_S(w_i) \leq 2M\delta_n + L \sqrt{\frac{2}{n}} + (D_S(w_i) - \Eof[S]{D_S(w_i)}).
    \end{align*}
    Thus:
    \begin{align*}
    \gnabla \leq \max_{1 \leq i \leq N_{\delta_n} } \left( D_S(w_i) - \Eof[S]{D_S(w_i)} \right) + L \sqrt{\frac{2}{n}} + 2M\delta_n.
    \end{align*}
    Using the independence between $\ycalu$ and a union bound, we get:
    \begin{align*}
    \mathds{P}_S \left( \max_{1 \leq i \leq N_{\delta_n} } (D_S(w_i) - \Eof[S]{D_S(w_i)}) \geq \epsilon \right) \leq N_{\delta_n} \exp \left\{ -\frac{n \epsilon^2}{2 L^2} \right\}.
    \end{align*}
    Now we set $\delta_n\defeq 1/M\sqrt{n}$. Using the same reasoning than in \cref{thm:lipschitz_master_theorem}, based on Egoroff's theorem applied on $\mu_u$, we that, for $n$ big enough (depending on $\zeta$), with probability at least $1 - 2\zeta$:
    \begin{align*}
    \gnabla \leq \frac{2}{\sqrt{n}} + L \sqrt{\frac{2}{n}} + 2L \sqrt{\frac{2\dimvalue \log(M\sqrt{n}) + \log(1/\zeta)}{2n}}.
    \end{align*}
    \textbf{With additional \cref{ass:partal_derivatives_concentration}}: We use the same covering and notations, but this time we write:
    \begin{align*}
        \gnabla \leq 2M\delta_n + \max_{1 \leq i \leq N_{\delta_n} } \sqrt{d \max_{1\leq j \leq d} \left| \frac{1}{n} \sum_{i=1}^n \partialmean(w,z_i) \right|^2}.
    \end{align*}
    From our additional assumption, we know that $\Vert \partialmean(w,z) \Vert_{\psi_2} = \Sigma/\sqrt{d}$ (where $\Vert \cdot \Vert_{\psi_2}$ is the sub-Gaussian norm), therefore by a union bound and Hoeffding's inequality:
    \begin{align*}
        \mathds{P}_S \left( \max_{1 \leq i \leq N_{\delta_n} }  \max_{1\leq j \leq d} \left| \frac{1}{n} \sum_{i=1}^n \partialmean(w,z_i) \right| \geq \epsilon \right) \leq 2dN_{\delta_n} \exp \left\{  
        -\frac{n d \epsilon^2}{2\Sigma^2} \right\}.
    \end{align*}
    Therefore, we set (remembering the $U$ dependence here):
    \begin{align*}
    \epsilon(U) \defeq \sqrt{2}\Sigma \sqrt{\frac{\log(N_{\delta_n}(U)) + \log(2d/\zeta)}{nd}}.
    \end{align*}
    We use again an argument based on Egoroff's theorem on $\mu_u$ to write that, for $N$ big enough (depending on $\zeta$), we have with probability $1 - \zeta$ that:
    \begin{align*}
    \gnabla \leq \frac{2}{\sqrt{n}} +  2 \Sigma \sqrt{\frac{2\dimvalue \log(M\sqrt{n}) + \log(2d/\zeta)}{2n}}.
    \end{align*}
\end{proof}

We are now ready to prove \cref{thm:second_main_result}.

\theoremBoundingHausdorff*

\begin{proof}
    We know from \cref{thm:heavy_tiled_sdes}, that the Feller processes $W^S$ and $Y$ have càdlàg paths (the variable $U$ is omitted).
    Thanks to \cref{sec:technical_lemmas}, we know that the Feller processes $W^S$ and $Y$ have $\mu_z^{\otimes n} \otimes \mu_u$-almost surely càdlàg paths and that the difference $V^S := W^S - Y$ is continuous (almost surely).
    Therefore, this continuity will be assumed in this proof without losing generality.
    As a consequence, the paths are almost surely bounded, so it makes sense to introduce their upper box-counting dimension.

    Let us take some $S\in\zcal^n$, we have, almost surely, for $t \leq T$:
    \begin{align*}
        \Vert V^S_t \Vert &= \normof{\int_0^t \left(  -\nabla \er(W_\tau^S) + \nabla\risk(Y_\tau) \right)  d\tau} \\
        &\leq\int_0^t \Vert \nabla \er(W_\tau^S) - \nabla\risk(Y_\tau) \Vert d\tau \\
        &\leq \int_0^t \Vert \nabla \er(W_\tau^S) - \nabla\er(Y_\tau) \Vert d\tau  + \int_0^t \Vert \nabla \er(Y_\tau) - \nabla\risk(Y_\tau) \Vert d\tau \\
        &\leq M \int_0^t  \Vert V^S_\tau \Vert d\tau + t \gnabla.
    \end{align*}
    From which we deduce:
    \begin{align*}
    \Vert V^S_t \Vert + \frac{1}{M} \gnabla \leq M& \int_0^t \left(  \Vert V^S_\tau \Vert + \frac{1}{M} \gnabla\right) d\tau \\&+ \frac{1}{M} \gnabla .
    \end{align*}
    By Grönwall's lemma, thanks to the continuity of $V^S_t$:
    \begin{align*}
    \Vert V^S_t \Vert \leq \frac{1}{M} \gnabla \left( e^{Mt} - 1 \right).
    \end{align*}
     The results immediately follow by \cref{lemma:bounding_gnabla} and rearranging terms.
\end{proof}

\begin{remark}
    The correct way of writing \cref{eq:empirical_dynamics,eq:expected_dynamics} would be:
    \begin{align*}
        dW^S_t = -\nabla \risk(W^S_{t-}) dt + \levy, \quad dY_t = -\nabla \risk(Y_{t-}) dt + \levy.
    \end{align*}
    It is clear, from the proof, that this detail does not create any issue, as the function $t \longmapsto V^S_t$ is still almost surely continuous.
\end{remark}

\subsection{Proof of our PAC-Bayesian Bounds}
\label{sec:proof_of_pac_bayes_bound}

In all this section, the notations $\pi_U$ and $\rho_{S,U}$ are the ones defined in \cref{sec:main-result-pac-bayes}, as well as the unperturbed posteriors and priors $\Tilde{\rho}_{S,U} $ and $\Tilde{\pi}_{U}$.
Let us also recall what we mean by the convolution of a measure with a function.
Let $\mu$ be a finite Borel measure on $\Rd$ and $f$ be a bounded measurable function, we define their convolution as:
\begin{align}
    \label{eq:convolution}
    f * \mu (x) = \int_{\Rd} f(x-y) d\mu(y).
\end{align}
When $f$ is a probability density, we will also denote by $f*\mu$ the associated probability measure.
We now quickly recall the definition of Wasserstein distance.

\begin{definition}[Wassertein distance]
    Let $\mu$ and $\nu$ be two probability measures on $\Rd$ and $p >0$. A \emph{coupling} between $\mu$ and $\nu$ is a probability measure on $\Rd \times \Rd$ whose marginals are $\mu$ and $\nu$. If $\Pi(\mu, \nu)$ denote the set of all possible such couplings, then we define the $p$-Wassertein distance by:
    \begin{align*}
    W_p(\mu, \nu) \defeq \inf_{\gamma \in \Pi(\mu, \nu)} \left\{ \iint \normof{x-y}^p d\gamma(x,y) \right\}^{\frac{1}{p}}.
    \end{align*}
\end{definition}

\begin{remark}[The PAC-Bayes analysis works in this non-classical setting]
Our framework may seem slightly non-standard, as the prior $\pi_U$ is random, because of its dependence on $U$.
However, we argue that, due to the independence of $\pi_U$ on $S$, the theorems from classical PAC-Bayesian theory, presented in \cref{sec:pac_bayes_background}, are still valid.
Moreover, one can see, from our measurability assumptions of \cref{rq:well-posedness} and \cref{sec:technical_lemmas}, that $\pi_U$ and $\rho_{S,U}$ are well-defined Markov kernels.
\end{remark}

\subsubsection{Bounds on Information-theoretic Distances}

In this subsection, we prove that the quantities $\klb{\rho_{S,U}}{\pi_U}$ and $\renyi{\rho_{S,U}}{\pi_U}$ can be bounded by the geometric distance.
The next lemma first takes care of the KL divergence, it is a simple application of Jensen's inequality, using the fact that the KL divergence is jointly convex in its argument \citep[Theorem $11$]{van_erven_renyi_2014}, we just carefully verify that it works in a continuous setting.

\begin{lemma}
    \label{lemma:KL_convolution_bound}
    With the previous notations, we have almost surely with respect to $S$ and $U$:
    \begin{align*}
    \klb{\rho_{S,U}}{\pi_U} \leq \frac{1}{2s^2} W_2(\Tilde{\rho}_{S,U},\Tilde{\pi}_U)^2 \leq \frac{1}{2s^2 T} \int_0^T \normof{W_t^S - Y_t}^2 dt.
    \end{align*}
\end{lemma}

The first inequality in this Theorem has already been used in \citep[Section $4.3$]{lugosi_generalization_2022}.
For the sake of clarity, we present a proof adapted to our framework.
As mentioned before, it is a consequence of the convexity of KL divergence.

\begin{proof}
    We start by proving a slightly more general result: let $\mu$ and $\nu$ be any probability measures on $\Rd$ and $f$ the density of $\mathcal{N}(0,s^2)$.     
    Let us define $F(x) = x \log(x)$. We shall note that $F$ is convex and bounded from below (this last point authorizes the subsequent use of Fubini's theorem in the proof). Let also $\gamma$ be any coupling between $\mu$ and $\nu$, we have, by Jensen's inequality:
    \begin{align*}
        F\left( \frac{d(f * \mu)}{d(f * \nu)}(x) \right) &= F\left( \int  \frac{f(x - y)}{f * \nu(x)}  d\mu(y) \right)\\
        &=  F\left( \iint \frac{f(x - y)}{f(x - z)} \frac{f(x - z)}{f * \nu(x)} d\gamma(y, z)  \right)\\
        &\leq \iint F\left(\frac{f(x - y)}{f(x - z)}\right) \frac{f(x - z)}{f * \nu(x)} d\gamma(y, z)
    \end{align*}
    Now we integrate over $f * \nu$ and apply Fubini's theorem (which we can do because $F$ is bounded from below and the measures are probability measures.):
    \begin{align*}
         \klb{f * \mu}{f * \nu} &\leq \iiint F\left(\frac{f(x - y)}{f(x - z)}\right) f(x - z) dx d\gamma(y, z) \\
         &= \iint \klb{\mathcal{N}(y, s^2)}{\mathcal{N}(z, s^2)} d\gamma(y, z) \\
         &= \frac{1}{2s^2} \iint \normof{y - z}^2 d\gamma(y, z),
    \end{align*}
    where we used the formula for KL divergence between Gaussians see \citep[Equation $10$]{van_erven_renyi_2014}.    
    Thus, by taking the infimum over all couplings:
    \begin{align}
        \label{eq:kl_convexity}
        \klb{f * \mu}{f * \nu}\leq \frac{1}{2s^2} W_2(\mu, \nu)^2.
    \end{align}
    Now we get back to $\rho_{S,U}$ and $\pi_U$, and as coupling, between $\Tilde{\rho}_{S,U}$ and $\Tilde{\pi}_U$, we chose the following
    \begin{align*}
    \gamma = (Y_\cdot, W^S_\cdot)_\# \mathcal{U}([0,T]),
    \end{align*}
    so that we have
    \begin{align*}
    W_2(\Tilde{\rho}_{S,U},\Tilde{\pi}_U)^2 \leq \frac{1}{T} \int_0^T \normof{W_t^S - Y_t}^2 dt.
    \end{align*}
    \cref{eq:kl_convexity} then implies the result.
\end{proof}

We now have the following corollary.

\begin{corollary}
    In the same setting, we have, almost surely, with respect to $S$ and $U$:
    \begin{align*}
     \klb{\rho_{S,U}}{\pi_U} \leq \sup_{0 \leq t \leq T} \normof{W_t^S - Y_t}^2.
    \end{align*}
\end{corollary}

The following lemma shows a similar bound for the Rényi divergence.
While \cref{lemma:KL_convolution_bound} has already been used in the literature, this next bound is new, to our knowledge.
It is slightly more complex because $\renyi[\beta]{\cdot}{\cdot}$ is not jointly convex for $\beta > 1$, the proof will actually extend the fact that it is jointly \emph{quasi}-convex \citep[Theorem $13$]{van_erven_renyi_2014}. 

\begin{lemma}
    \label{lemma:renyi_convolution_bound}
    Let us fix $\beta > 1$, with the same notations as before, we have:
    \begin{align*}
    \renyi[\beta]{\rho_{S,U}}{\pi_U} \leq \frac{\beta}{2s^2} \sup_{0 \leq t \leq T} \normof{W_t^S - Y_t}^2.
    \end{align*}
\end{lemma}

\begin{proof}
    We start with a similar calculation to the proof of \cref{lemma:KL_convolution_bound}.
    Let $\mu$ and $\nu$ be any probability measures on $\Rd$ and $f$ the density of $\mathcal{N}(0,s^2)$.
    Let also $\gamma$ be a coupling between $\mu$ and $\nu$.
    By Jensen's inequality, we have:
    \begin{align*}
        \left\{  \frac{d(f * \mu)}{d(f * \nu)}(x) \right\}^\beta &=  \left\{  \iint  \frac{f(x - y)}{f(x - z)} \frac{f(x - z)}{(f * \nu(x))} d\gamma(y, z) 
         \right\}^\beta \\
         &\leq  \iint  \left(\frac{f(x - y)}{f(x - z)}\right)^\beta \frac{f(x - z)}{f * \nu(x)} d\gamma(y, z) \\
         &= \frac{1}{f * \nu(x)} \iint f(x - y)^\beta f(x - z)^{1 - \beta} d\gamma(y, z),
    \end{align*}
    where the inequality holds by Jensen's inequality because we have
    \begin{align*}
         \iint  \frac{f(x - z)}{(f * \nu(x))} d\gamma(y, z) = \int  \frac{f(x - z)}{(f * \nu(x))} d\nu(z) = 1.
    \end{align*}
    Now we integrate with respect to $f * \nu$ and take the logarithm, by Fubini's theorem (all functions are positive), it follows that:
    \begin{align*}
         \log \left( \int (f * \mu(x))^\beta (f * \nu(x))^{1 - \beta} dx\right) &\leq \log \left( \iiint f(x - y)^\beta f(x - z)^{1 - \beta} dx d\gamma(y, z)\right)\\
         &\leq \log \left( \esssup_{y, z \sim \gamma} \int f(x - y)^\beta f(x - z)^{1 - \beta} dx \right)
    \end{align*}
    By dividing by $\beta - 1$ and putting the $\log$ inside the essential supremum, we get:
    \begin{align*}
    \renyi[\beta]{\mu}{\nu} &\leq \log \left( \esssup_{y, z \sim \gamma} \frac{1}{\beta-1}\int f(x - y)^\beta f(x - z)^{1 - \beta} dx \right) \\
    &= \esssup_{y, z \sim \gamma}\left(  \renyi[\beta]{\mathcal{N}(y,s^2)}{\mathcal{N}(z,s^2)} \right) \\
    &=   \esssup_{y, z \sim \gamma}\left( \frac{\beta}{2s^2} \normof{y - z}^2 \right),
    \end{align*}
    where we used known formula, deduced from \citep[Equation $10$ and Theorem $28$]{van_erven_renyi_2014}, to compute the divergence between Gaussians.
    Now, we get back to $\rho_{S,U}$ and $\pi_U$, and as coupling, between $\Tilde{\rho}_{S,U}$ and $\Tilde{\pi}_U$, we chose the same as in the proof of the previous lemma:
    \begin{align*}
    \gamma = (Y_\cdot, W^S_\cdot)_\# \mathcal{U}([0,T]),
    \end{align*}
    so that we have
    \begin{align*}
    \renyi{\rho_{S,U}}{\pi_U} \leq \sup_{0 \leq t \leq T} \normof{W_t^S - Y_t}^2.
    \end{align*}
\end{proof}

\subsubsection[Proof of Th. 5]{Proof of \cref{thm:pac_bayes_wassertein}}

\theoremPACBayesWassertein*

The random variable $U$ is implicitly omitted when proving those bounds, by independence, it does not change the results (all the results may be true $U$-almost surely).
Note that, to simplify notations, we denote $\mathds{E}_{\rho_{S,U}}$ for $\mathds{E}_{w \sim \rho_{S,U}}$ (and idem for $\pi_U$).

\begin{proof}
    Let $\lambda > 0$ and $\zeta \in (0,1)$. We simply write the result of \cref{thm:kl_pac_bayes__bound} with the previously defined priors and posteriors, and with function:
    \begin{align*}
        \Phi(w,S) \defeq \lambda \left( \risk(w) - \er(w) \right).
    \end{align*}
    This gives with probability at least $1-\zeta$ over $S\sim\mu_z^{\otimes n}$
    \begin{align*}
        \lambda  \Eof[\rho_{S,U}]{ \risk(w) - \er(w)} \leq \log(1/\zeta) + \klb{\rho_{S,U}}{\pi_U} + \log \mathds{E}_S \Eof[\pi_U]{e^{\lambda \left( \risk(w) - \er(w) \right)}}.
    \end{align*}
    The last term can be bounded classically, using the sub-Gaussian assumption and Fubini's theorem:
    \begin{align*}
        \mathds{E}_S \Eof[\pi]{e^{\lambda \left( \risk(w) - \er(w) \right)}} &= \mathds{E}_{\pi_U} \Eof[S]{e^{\lambda \left( \risk(w) - \er(w) \right)}} \\
        &\leq \exp \left\{ \frac{ \lambda^2 \sigma^2}{2n} \right\}.
    \end{align*}
    By \cref{lemma:KL_convolution_bound}, we have:
    \begin{align*}
     \klb{\rho_{S,U}}{\pi_U} \leq \frac{1}{2s^2} W_2(\bar{\pi_U}, \bar{\rho}_S)^2 \leq \frac{1}{2s^2 T} \int_0^T \normof{W_t^S - Y_t} dt.
    \end{align*}
    The result follows.
\end{proof}

\subsubsection[Proof of Th. 6]{Proof of \cref{thm:disintegrated_heavy_tailed_bound}}

\thmDisintegrated*

\begin{proof}
    It is exactly the same thing than for \cref{thm:pac_bayes_wassertein}, but instead of \cref{thm:kl_pac_bayes__bound}, we use \cref{thm:disintegrated_pac_bayes__bound}, and instead of \cref{lemma:KL_convolution_bound}, we use \cref{lemma:renyi_convolution_bound}.
\end{proof}

\subsection[Proof of Th. 13]{Proof of \cref{thm:co_dissipativity_bound_on_sup}}

\thmCoDissipativity*

\begin{proof}
    The intuition of the proof is as follows: to use \cref{ass:co_dissipativity}, we would like to differentiate the function $t \longmapsto \normof{W_t^S - Y_t}^2$ and then use the differential form of Grönwall Lemma, along with estimates on the quantity $\gnabla$ introduced in \cref{lemma:bounding_gnabla}. However, the aforementioned function is clearly non differentiable everywhere\footnote{But we can see, from Rademacher theorem, that it is differentiable almost everywhere.}, because of the jumps of the driving Lévy process. While we could chain the above argument in case the number of jumps is almost surely finite (it is the case, for example, if $\alpha=2$), we propose here another argument, based on the regularization of the function by convolution. Let $V^S = W^S - Y$, which we continuously extend to $\R$ by setting the value $0$ on $(-\infty, 0)$. We omit the dependence on $U$.

    Before, describing the rest of the proof, let us take $w \in \Rd$, and $\mathcal{U}$ a compactly contained open set that contains $w$. Then we have, by the smoothness assumption:
    \begin{align*}
        \normof{\nabla \ell(w,z)} \leq \normof{\nabla \ell(w_0,z)} + M \sup_{w \in \mathcal{U}} \normof{w - w_0},
    \end{align*}
    which ensures, by the dominated convergence theorem, that we can write $\nabla \risk(w) = \Eof[S]{\nabla \ell(w,z)}$, without issues.

    For the rest of the proof, we fix $S \in\zcal^n$, $U \in \Omega$ as well as $T>0$. 

    Before all, note that, even without the Lipschitz assumption, we can show that the paths of the processes $W^S$ and $Y$ are almost surely bounded, see Remark \ref{rq:not_lipschitz_not_feller}, as they admit cadlag modifications.
    Therefore, we can assume, without loss of generality, that, for all $T$, there exists a compact $K_T \subset \Rd$ such that, for all $0\leq t \leq T$, $W_t^S, y_t \in K_T$.

    Let $(\varphi)_{\epsilon>0}$ be a regularizing sequence, \ie a sequence of functions $\varphi_\epsilon:\R \longrightarrow \R$ such that:
    \begin{align*}
        \varphi_\epsilon \in \mathcal{C}^\infty, \quad \varphi_\epsilon \geq 0, \quad \support (\varphi_\epsilon) \subset (-\epsilon, \epsilon), \quad \intr \varphi_\epsilon(x) dx = 1.
    \end{align*}
    First note that we have:
    \begin{align*}
        \varphi_\epsilon * V^S_t &= \intr \phieps'(t - s) V^S_s ds \\
                &= \intr \int_0^s \phieps'(t - s) \left( -\nabla \er(W^S_u) + \nabla \risk(Y_u) \right) du ds \\
                &= \intr \left\{ \int_s^\infty \phieps'(t - s) ds \right\} \left( -\nabla \er(W^S_u) + \nabla \risk(Y_u) \right) du \\
                &= \intr  \phieps(t - s) \left( -\nabla \er(W^S_u) + \nabla \risk(Y_u) \right) du \\
    \end{align*}
    now we compute that, as, by properties of convolution, the derivative of $\phieps * V_t^S$ is $\phieps * V_t^S$:
    \begin{align*}
        \frac{d}{dt} \Vert \phieps * V_t^S \Vert^2 &= 2 \langle \phieps' * V_t^S, \phieps * V_t^S \rangle \\
        &= 2 \langle \phieps * \left( -\nabla \er(W^S_\cdot) + \nabla \risk(Y_\cdot) \right)(t), \phieps * V_t^S \rangle \\
        &=: 2(A+B),
    \end{align*}
    where, for $t\in [0,T]$ and some fixed arbitrary $\eta > 0$, for $\epsilon < \eta$:
    \begin{align*}
        B &\defeq \langle \phieps * \left( -\nabla \er(Y_\cdot) + \nabla \risk(Y_\cdot) \right)(t), \phieps * V_t^S \rangle \\
        &\leq \Vert  \phieps * V_t^S \Vert \cdot \left\Vert \intr  \phieps(t - s) \left( -\nabla \er(W^S_u) + \nabla \risk(Y_u) \right) du \right\Vert \\
        &\leq \Vert  \phieps * V_t^S \Vert G_\nabla^\eta \intr \phieps(t-u) du \\
        &= \Vert  \phieps * V_t^S \Vert G_\nabla^\eta,
    \end{align*}
    where $G_\nabla^{\eta}$ is like in \cref{eq:gnabla}, but on $[0,T+\eta]$.
    \begin{align*}
    G_\nabla^{\eta} \defeq  \sup_{0 \leq t \leq t+\eta} \Vert \nabla\er(Y_t) - \nabla \risk(Y_t) \Vert.
    \end{align*}
    Now we have:
    \begin{align*}
        A &\defeq \intr  \intr \phieps(t-s) \phieps(t-u) \langle -\nabla \er(W^S_s) + \nabla \er(Y_s), W^S_u - Y_u \rangle ds du \\
        &=: C+D,
    \end{align*}
    where, by \cref{ass:co_dissipativity}:
    \begin{align*}
        C &\defeq \intr  \intr \phieps(t-s) \phieps(t-u) \langle -\nabla \er(W^S_s) + \nabla \er(Y_s), W^S_s - Y_s \rangle ds du \\
        &\leq   \intr \phieps(t-s) \left( -m\Vert W_s^S - Y_s \Vert^2 + K \right) ds  \\
        &= -m \intr \phieps(t-s)\Vert W_s^S - Y_s \Vert^2  ds + K 
    \end{align*}
    By Cauchy-Schwarz's inequality, we have:
    \begin{align*}
        \intr \phieps(t-s)\Vert W_s^S - Y_s \Vert^2  ds \intr \phieps(t-s) ds \geq \left\{ \intr \phieps(t-s)\Vert W_s^S - Y_s \Vert ds \right\}^2,
    \end{align*}
    so that by the triangle inequality:
    \begin{align*}
        C \leq -m  \Vert \phieps * V_t^S \Vert^2 + K .
    \end{align*}
    Let us introduce, thanks to the continuity of the functions $\nabla \er$ and $\nabla \risk$ and the fact that the trajectory is in the compact $K_{T+\eta}$:
    \begin{align*}
    G := \max \left\{\sup_{x \in K_{T+\eta}} \Vert \nabla \risk(x) \Vert, \sup_{x \in K_{T+\eta}} \Vert \nabla \er(x) \Vert \right\}
    \end{align*}
    On the other hand, by noting that $W_\cdot^S - Y_\cdot$ is $2L$-Lipschitz continuous on $\R$ and using Cauchy-Schwarz's inequality, for $\epsilon < \eta$:
    \begin{align*}
        D &\defeq  \intr  \intr \phieps(t-s) \phieps(t-u) \langle -\nabla \er(W^S_s) + \nabla \er(Y_s), W^S_u - Y_u - W^S_s + Y_s \rangle ds du \\
        &\leq 2G \intr  \intr \phieps(t-s) \phieps(t-u) \Vert -\nabla \er(W^S_s) + \nabla \er(Y_s) \Vert \cdot |s - u| ds du\\
        &= 4 G^2 \int_{t - \epsilon}^{t + \epsilon} \int_{t - \epsilon}^{t + \epsilon}  \phieps(t-s) \phieps(t-u) |s - u| ds du\\
        &\leq 8 G^2 \epsilon.
    \end{align*}
    Therefore, we have deduced that, for $t \in [0,T]$:
    \begin{align*}
         \frac{d}{dt} \Vert \phieps * V_t^S \Vert^2 \leq 2 \Vert  \phieps * V_t^S \Vert G_\nabla^\eta -2m  \Vert \phieps * V_t^S \Vert^2 + 2K + 16G^2\epsilon.
    \end{align*}
    By Young's inequality, we further have:
    \begin{align*}
         \frac{d}{dt} \Vert \phieps * V_t^S \Vert^2 \leq \frac{(G_\nabla^\eta)^2}{m} -m  \Vert \phieps * V_t^S \Vert^2 + 2K + 16G^2\epsilon.
    \end{align*}
    Thus, by Gronwall's Lemma:
    \begin{align*}
        \Vert \phieps * V_t^S \Vert^2  \leq \frac{(G_\nabla^\eta)^2 + 2Km + 16 G^2 m \epsilon}{m^2} (1 - e^{-mt}) +  \Vert \phieps * V_t^S \Vert^2 \bigg|_{t=0} e^{-mt}.
    \end{align*}
    Now, as $t \longmapsto V_t^S$ is uniformly continuous (because it is Lipschitz continuous, thanks to the boundedness of $\nabla_w \ell$ on the trajectories of $W^S$ and $Y$, which are included in a compact), see \cref{sec:technical_lemmas}, we know that $\phieps * V_t^S$ converges uniformly to $V_t^S$ on $[0,T]$, hence, by taking the limit $\epsilon \to 0$, we get, for $t \in [0,T]$:
    \begin{align*}
        \Vert  V_t^S \Vert^2 \leq  \frac{(G_\nabla^\eta)^2 + 2Km }{m^2} (1 - e^{-mt}) 
    \end{align*}
    As $\eta$ is arbitrary, we can now simply apply the same reasoning as in the proof of \cref{lemma:bounding_gnabla}. It gives us that, for $n$ large enough, with probability at least $1 - \zeta$ over $\mu_u \otimes \mu_z^{\otimes n}$, we have:
    \begin{align*}
    \gnabla \leq \frac{2}{\sqrt{n}} +  2 \Sigma \sqrt{\frac{2\dimvalue \log(M\sqrt{n}) + \log(2d/\zeta)}{2n}}.
    \end{align*}
    So finally, for $n$ large enough, with probability at least $1 - \zeta$ over $\mu_u \otimes \mu_z^{\otimes n}$, we have:
    \begin{align*}
        \sup_{0 \leq t \leq T}  \Vert V_t^S \Vert^2 \leq \frac{2K}{m} + \frac{4}{m^2} \bigg\{ \frac{2}{n} + \frac{\Sigma^2}{n} \left( 2\dimvalue \log(M\sqrt{n}) + \log(2d/\zeta) \right) \bigg\}.
    \end{align*}
\end{proof}

\begin{remark}
    Plugging the above bound in the PAC-Bayesian bounds presented before, with $\lambda = \sqrt{n}$, will give a higher convergence rate in $n$ to the terms having big constants.
    Moreover, the constant $\frac{K}{2m}$ term is not a problem, as it will be of the order $\frac{K}{2m\sqrt{n}}$ in the PAC-Bayesian bounds.
\end{remark}

\end{document}